%% file: main.tex
\pdfoutput=1

\documentclass[11pt]{article}

\usepackage[final]{acl}

\usepackage{times}
\usepackage{latexsym}
\usepackage{hyperref}
\usepackage[T1]{fontenc}

\usepackage[utf8]{inputenc}

\usepackage{microtype}

\usepackage{inconsolata}

\usepackage{booktabs}
\usepackage{multirow}
\usepackage{tikz}
\usepackage{amsmath,amssymb}
\usepackage{float}     
\usepackage{placeins}  
\usepackage{enumitem}
\usepackage{graphicx}
\usepackage{tikz}
\usepackage{mathtools}
\usepackage{comment}
\usepackage{makecell}
\usepackage{subcaption}
\usepackage[capitalize,nameinlink]{cleveref}
\DeclarePairedDelimiter{\ceil}{\lceil}{\rceil}
\newcommand{\spacehack}[1]{\relax}

\definecolor{nodefill}{RGB}{218,232,252}
\definecolor{nodedraw}{RGB}{108,142,191}
\newcommand*\circled[1]{\tikz[baseline=(char.base)]{
            \node[fill=nodefill,shape=circle,draw=nodedraw,inner sep=0.3pt] (char) {#1};}
            }

\definecolor{nodefillred}{RGB}{	248, 206, 204}
\definecolor{nodedrawred}{RGB}{184, 84, 80}
\newcommand*\circledred[1]{\tikz[baseline=(char.base)]{
            \node[fill=nodefillred,shape=circle,draw=nodedrawred,inner sep=0.3pt] (char) {#1};}
            }

\newcommand{\ie}{\textit{i.e.}}
\newcommand{\eg}{\textit{e.g.}}

\makeatletter

\providecommand{\section}{}
\renewcommand{\section}{%
  \@startsection{section}{1}{\z@}%
                {-1.5ex \@plus -0.5ex \@minus -0.2ex}%
                { 1.0ex \@plus  0.3ex \@minus  0.2ex}%
                {\large\bf\raggedright}%
}
\providecommand{\subsection}{}
\renewcommand{\subsection}{%
  \@startsection{subsection}{2}{\z@}%
                {-1.3ex \@plus -0.5ex \@minus -0.2ex}%
                { 0.3ex \@plus  0.2ex}%
                {\normalsize\bf\raggedright}%
}
\providecommand{\subsubsection}{}
\renewcommand{\subsubsection}{%
  \@startsection{subsubsection}{3}{\z@}%
                {-1.0ex \@plus -0.5ex \@minus -0.2ex}%
                { 0.3ex \@plus  0.2ex}%
                {\normalsize\bf\raggedright}%
}
\providecommand{\paragraph}{}
\renewcommand{\paragraph}{%
  \@startsection{paragraph}{4}{\z@}%
                {1.0ex \@plus 0.5ex \@minus 0.2ex}%
                {-1em}%
                {\normalsize\bf}%
}
\providecommand{\subparagraph}{}
\renewcommand{\subparagraph}{%
  \@startsection{subparagraph}{5}{\z@}%
                {1.5ex \@plus 0.5ex \@minus 0.2ex}%
                {-1em}%
                {\normalsize\bf}%
}

\makeatother

\title{Optimized Speculative Sampling for GPU Hardware Accelerators}

\author{
 \textbf{Dominik Wagner\textsuperscript{1}},
 \textbf{Seanie Lee\textsuperscript{2}},
 \textbf{Ilja Baumann\textsuperscript{1}},
 \textbf{Philipp Seeberger\textsuperscript{1}},
\\
 \textbf{Korbinian Riedhammer\textsuperscript{1}},
 \textbf{Tobias Bocklet\textsuperscript{1}}
\\
 \textsuperscript{1}Technische Hochschule Nürnberg Georg Simon Ohm,
 \textsuperscript{2}KAIST
\\
\texttt{dominik.wagner@th-nuernberg.de} \\
\small\texttt{lsnfamily02@kaist.ac.kr}\\
\small$\texttt{\{ilja.baumann, philipp.seeberger, korbinian.riedhammer, tobias.bocklet\}@th-nuernberg.de}$
}

\begin{document}
\maketitle
\begin{abstract}
In this work, we optimize speculative sampling for parallel hardware accelerators to improve sampling speed. 
We notice that substantial portions of the intermediate matrices necessary for speculative sampling can be computed concurrently. 
This allows us to distribute the workload across multiple GPU threads, enabling simultaneous operations on matrix segments within thread blocks. 
This results in \emph{profiling time improvements} ranging from 6\% to 13\% relative to the baseline implementation, \emph{without compromising accuracy}. 
To further accelerate speculative sampling, probability distributions parameterized by softmax are approximated by sigmoid. 
This approximation approach results in significantly greater relative improvements in profiling time, ranging from 37\% to 94\%, with a  minor decline in accuracy.
We conduct extensive experiments on both automatic speech recognition and summarization tasks to validate the effectiveness of our optimization methods.
\end{abstract}
\input{1_intro}
\input{2_related_work}
\input{3_method}
\input{4_exp}
\input{5_conclusion}
\input{6_limitation}

\section*{Acknowledgments}
We gratefully acknowledge the scientific support and HPC resources provided by the Erlangen National High Performance Computing Center (NHR@FAU) of the Friedrich-Alexander-Universität Erlangen-Nürnberg (FAU) under the NHR project b196ac14. NHR funding is provided by federal and Bavarian state authorities. NHR@FAU hardware is partially funded by the German Research Foundation (DFG) – 440719683.
This work was supported by the Bavarian State Ministry of Science and the Arts under grant H.2-F1116.NÜ/61/2. 

\bibliography{refs}

\clearpage
\appendix

\input{7_appendix}

\end{document}

%% file: 1_intro.tex
\section{Introduction}\label{sec:intro}
Large foundational speech and language models  based on autoregressive Transformer~\cite{vaswani17attention} architectures have demonstrated remarkable proficiency across a variety of downstream tasks~\cite{weining2021hubert,radford2022whisper,touvron2023llama2,openai2024gpt4}. 
These models frequently increase in size, consequently requiring more memory and computational resources.
However, downstream applications, such as dialogue systems, have strict wall-clock constraints and are often required to generate long sequences~\cite{pope2022efficiently,chi2023distillery,fischer2024grillbot}. 
Due to the sequential token generation in autoregressive decoding, latency increases with both the length of the sequence and the size of the model, resulting in a significant barrier to widespread deployment. 
On many general-purpose GPU hardware accelerator architectures, the increasing size of models leads to more read and write operations between high-bandwidth memory (HBM) and on-chip shared memory (SRAM) at each decoding step, necessitating more memory bandwidth to meet latency constraints~\cite{pope2022efficiently,dao2022flashattention,dao2024flashattention2}.
Consequently, the speed of autoregressive decoding becomes primarily limited by the available memory bandwidth and not by the number of computations that need to be executed on the dedicated hardware \cite{shazeer2019fast}. 

In many cases, however, tokens may be accurately generated by much smaller models that require fewer resources. 
Motivated by this hypothesis, speculative sampling has been developed to accelerate autoregressive sampling~\cite{stern2018specdec,xia-etal-2023-speculative,leviathan2023decoding,chen2023sampling}. 
Speculative sampling employs a small draft model to generate tokens, which are potential future outputs of a larger target model.
These drafted tokens are then verified in parallel by the target model, and only tokens that meet the validation criteria are retained as final outputs to ensure generation accuracy. 
This approach has been shown to significantly reduce the frequency of time-consuming operations, thereby improving inference latency \cite{leviathan2023decoding,chen2023sampling}.

In this paper, we focus on optimizing the validation part of speculative sampling to further increase the inference speed. 
Inspired by recent advances in accelerating computations in the attention mechanism \cite{dao2022flashattention,dao2024flashattention2}, we explore two faster methods for assessing drafted tokens by leveraging the parallelism capabilities of modern GPUs. 
We identify that a significant portion of the intermediate matrices required for the sampling process can be computed independently. 
Exploiting this observation, we distribute the workload across multiple GPU threads and simultaneously compute portions of the intermediate output matrices within thread blocks. 
This optimization method is faster than the non-optimized baseline implementation and \textit{exact} with regard to the decoding outputs, \ie, it generates the same outputs as the non-optimized method.

To further accelerate speculative sampling, we propose using sigmoid as an element-wise approximation to softmax~\citep{bridle89softmax}, which is used to parameterize distributions of target and draft models.
Since sigmoid is applied to logits in element-wise fashion, it can be computed in parallel and fused with other sampling-related computations. This enables significant acceleration of the overall process, but results in a small accuracy decline due to the \textit{non-exact} nature of the method.

We evaluate our two optimized algorithms on automatic speech recognition (ASR) and summarization tasks, covering draft model sizes between 166M and 2B parameters and target model sizes between 244M and 13B parameters. 
The \textit{exact} optimization method reduces profiling time between 6\% and 13\% relative to the baseline implementation without compromising accuracy. 
Moreover, the \textit{non-exact} optimization method improves profiling time by 37\% to 94\%, albeit with a small reduction in accuracy.
We summarize our main contributions as follows:\footnote{The source code of our optimized sampling algorithm is available at \url{https://github.com/dwgnr/optimized-speculative-sampling}.} 

\begin{itemize}[itemsep=1mm,parsep=1pt,topsep=0pt,leftmargin=*]
    \item Implementation of an \textit{exact} and consistently faster variant of speculative sampling optimized for GPU hardware accelerators. 
    \item Exploration of sigmoid as an element-wise approximation to softmax in an even faster but \textit{non-exact} variant of speculative sampling. 
    \item Comprehensive evaluation across multiple tasks, covering a wide range of draft and target model combinations.
\end{itemize}

%% file: 2_related_work.tex
\section{Related work}\label{sec:related_work}

Techniques such as quantization
~\cite{dettmers2022gptint,bondarenko2023quantizable,stock2021training,nagel2021white}, pruning \cite{voita2019analyzing,lagunas2021blockpruning,gromov2024unreasonable} and knowledge distillation \cite{sun2019patient,sanh2020distilbert,jiao2020tinybert,hsieh-etal-2023-distilling} have proven effective in reducing inference latency with minimal performance impact. 
However, these approaches often require architectural changes or custom training procedures. 
Efforts specifically targeting the reduction of memory bandwidth bottlenecks during decoding include methods like multi-query attention, which aims to optimize memory usage per attention layer \cite{shazeer2019fast}, or FlashAttention~\citep{dao2022flashattention,dao2024flashattention2}, which aims to reduce the number of read/write operations between HBM and SRAM on GPUs. \citet{pope2022efficiently}, achieve improvements in large-scale inference latency by partitioning models and workload across multiple accelerators combined with various low-level optimizations to improve communication efficiency between devices. 

Speculative sampling approaches can be broadly categorized based on how drafting and verification are conducted \cite{xia2024unlocking}. 
\textit{Drafting} refers to the efficient prediction of multiple future tokens with a draft model and \textit{verification} refers to the methods used to verify the token sequence with the target model. 
Some works use specialized draft models \cite{xia-etal-2023-speculative,zhou2024distillspec}. 
Others employ an existing smaller model from the same series \cite{chen2023sampling,spector2023accelerating,leviathan2023decoding,yang2024multicandidate} or leverage the target model directly for drafting, \eg, by skipping intermediate layers \cite{zhang2023draft}, using special look-ahead tokens \cite{monea2023pass}, or additional modeling heads \cite{stern2018specdec,cai2024medusa,zhang2024recurrent}. 
Verification approaches first supported greedy decoding \cite{stern2018specdec,xia-etal-2023-speculative,zhang2023draft} and were subsequently extended to support other methods such as nucleus sampling \cite{leviathan2023decoding,chen2023sampling}. 
Recently, methods to verify multiple draft sequences in parallel have also been explored \cite{miao2024tokentree,cai2024medusa,spector2023accelerating}.

Several studies have experimented with ReLU and sigmoid as alternatives to softmax in the attention mechanism~\citep{bai2023transformers,shen2023study,Hron2020InfiniteAN,hua2022quality,li2022robust,wortsman2023replacing,ramapuram2024sigmoidatt}. 
They either require training new models from scratch or maintain the computational overhead of gathering information along the full sequence length. 
Other softmax-related optimizations are tailored to reduce the memory requirement during training by computing only smaller fractions of the full softmax output in the backward pass \cite{lee2023softmax} or leverage word frequencies in the training data to speed up computation \cite{grave2017adaptivesoftmax}. 
\citet{shim2017svdsoftmax} approximate softmax by computing only a fraction of the full input with singular value decomposition. 
Other approximation methods are specifically designed for custom hardware such as field-programmable gate arrays~\cite{ke2023approx} and application specific integrated circuits~\cite{Geng2018HardwareAwareSA}. 

%% file: 3_method.tex
\section{Method}\label{sec:method}
\subsection{Preliminaries}

\paragraph{Speculative sampling.}
Let $M_p^{\texttt{target}}$ be an autoregressive target model, which induces a categorical distribution distribution $p(x|x_{<i+1})$ over a vocabulary $\mathcal{V}=\{x\in\mathbb{N}: 1\leq x \leq \texttt{vocab\_size} \}$, given the prefix $x_{<i+1}=(x_1, \ldots,  x_{i})$. 
Our goal is to use speculative sampling to accelerate the sampling process of discrete tokens $x\in\mathcal{V}$. 
This is achieved by approximating the target model with a draft model $M_q^{\texttt{draft}}$, resulting in another categorical distribution $q(x|x_{<i+1})$.

First, given the prefix $(x_1,\ldots, x_{i+c-1})$, $\gamma\in\mathbb{N}_+$ draft tokens are sequentially sampled with $M_q^{\texttt{draft}}$: $x_{i+c}\sim q(x|x_{<i+c})$ for $c=1,\ldots, \gamma$. 
The draft tokens are then evaluated using the target model $M_p^{\texttt{target}}$, a process that can be performed in parallel. 
Each draft token $x_{i+c}\in\mathcal{V}$ is accepted if $r_c\leq \tau_{c}(x_{i+c})$ for $c=1,\ldots, \gamma$. 
The terms $r_{c}$ and $\tau_c(x_{i+c})$ are computed as follows:
\begin{equation}
\begin{gathered}
    \tau_{c}(x_{i+c})= \min \left(1, \frac{p({x}_{i+c}| {x}_{<i+c})}{q({x}_{i+c}| {x}_{<i+c})} \right) \\ r_c\sim U(0,1),
\end{gathered}
\label{eq:rejection}
\end{equation}
where $U(0,1)$ denotes a uniform distribution between 0 and 1.
If ${{x}}_{i+c}$ is accepted, the process is repeated for the next token $ {{x}}_{i+c+1}$ until either a token is rejected or all tokens have been accepted. 
If ${{x}}_{i+c}$ is rejected, a token is resampled from the following adjusted distribution instead: 
\begin{equation}
\begin{gathered}
{x}_{i+c} \sim \texttt{max\_norm}\left(p({x}|{x}_{<i+c}) - q({x}|{x}_{<i+c})\right),
\end{gathered}
\label{eq:modified}
\end{equation}
where $\texttt{max\_norm}(f(x))$ is given by:
\begin{equation}
\begin{aligned}
\texttt{max\_norm} (f(x))  &=  \frac{\max(0, f(x))}{{\sum_{x^\prime\in\mathcal{V}} \max(0, f(x^\prime))}} \\
                           &= \frac{a(x)}{b}
\end{aligned}
\label{eq:max-norm}
\end{equation}
We denote the numerator of~\autoref{eq:max-norm} by $a(x)$ and the denominator by $b$, as these terms are treated separately in subsequent sections. 

The underlying concept of speculative sampling is similar to rejection sampling~\citep{neal2003rejectionsampling}. 
Intuitively, a new token for the target model $M_p^{\texttt{target}}$ is generated by first sampling from a smaller draft model $M_q^{\texttt{draft}}$, which shares the same support ($\mathcal{V}$) as $M_p^{\texttt{target}}$.
The token sampled from $M_q^{\texttt{draft}}$ is then evaluated in parallel with $M_p^{\texttt{target}}$, and its acceptance is determined based on the probability ratio defined in~\autoref{eq:rejection}. 
If the token is rejected, a new one is drawn using the modified distribution in~\autoref{eq:modified}. 
\paragraph{GPU memory and execution model.}
We briefly describe the memory components and parts of the execution model of GPU hardware accelerators relevant to this work. 
GPU memory has a hierarchical layout, consisting of various types of memory that differ in size and read/write bandwidth~\citep{jia2018dissecting}. 
Recent GPUs (\eg, NVIDIA's A100 series) typically feature several gigabytes of high-bandwidth memory (HBM) and only a few hundred kilobytes of on-chip shared memory (SRAM) per streaming multiprocessor (SM) \cite{nvidia_a100_2020}. 
While HBM provides substantial capacity, its memory bandwidth is lower compared to SRAM. 
The execution model of GPU hardware accelerators involves a large number of threads executing operations known as kernels~\cite{cheng2014professional}. 
These threads are organized into thread blocks and assigned to SMs. 
Each SM partitions its assigned thread blocks into warps of 32 threads, which are then queued for execution on available hardware resources.
Each kernel typically follows a pattern: loading inputs from HBM into registers and SRAM, performing computations, and writing the outputs back to HBM. 
\subsection{Acceleration of speculative sampling}
\begin{figure*}[ht]
  \centering
\includegraphics[width=1.0\linewidth]{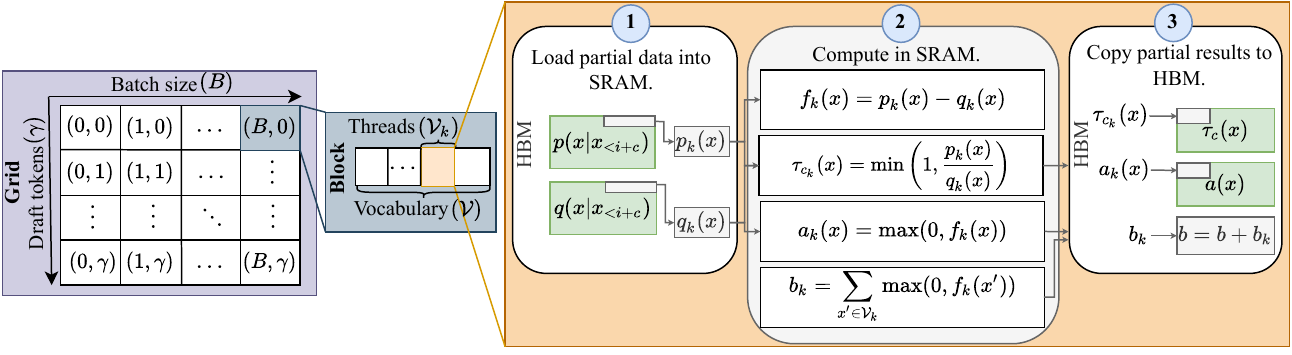}
  \vspace{-8mm}
  \caption{Overview of our exact optimization approach. We compute most of the results required for speculative sampling in parallel using fast SRAM to read and write intermediate results. We maximize the number of threads per block to run parallel computation on as many elements as possible without exhausting the available on-chip memory. 
  }
  \label{fig:approach}
  \vspace{-0.15in}
\end{figure*}
\subsubsection{Exact optimization}\label{ssec:exact}
Our optimization of speculative sampling is designed for parallel heterogeneous hardware accelerators, such as NVIDIA GPUs, which are widely employed to perform inference on large scale models. 
Similar to the approaches described in~\citet{ryoo2008optimization} and~\citet{dao2022flashattention}, we aim to redistribute the speculative sampling workload across threads and thread blocks. 
We load chunks of inputs from HBM to SRAM and make the necessary computations with respect to this input chunk before writing the final result back to HBM. 

We notice that the intermediate elements needed for speculative sampling can be computed concurrently within thread blocks and are largely independent of other thread blocks. 
Specifically, we can compute $(\tau_{c}(x))_{x\in\mathcal{V}}$ and parts of \autoref{eq:max-norm} in parallel.
The kernel is tiled~\cite{lam1991tiling}, such that each thread block computes a tile of $n$ elements from each matrix at once.
Threads within a block jointly load segments of the probability matrices $p(x|x_{x<i+c})$ and $q(x|x_{<i+c})$, whose full dimensions are $B\times \gamma\times\lvert\mathcal{V}\rvert$, into SRAM, effectively distributing the overhead of load latency \cite{ryoo2008optimization}. 
$B$ denotes the batch size used during speculative decoding. 

\autoref{fig:approach} illustrates the details of our approach. 
The overall workload is distributed across a two-dimensional grid of batch size $B$ and number of draft tokens $\gamma$. The vocabulary $\mathcal{V}$ is partitioned into disjoint subsets $\{\mathcal{V}_k \}_{k=1}^K$, where $K=\ceil{\lvert\mathcal{V}\rvert / n}$ and  $\ceil{\cdot}$ is the ceiling function. 
Each thread in a block performs operations on the sub-vocabulary $\mathcal{V}_k$ within its corresponding tile.
The results in each tile $k$ are obtained in three steps. \circled{\textbf{1}}The domains of the probability density functions $p(x|x_{i+c})$ and $q(x|x_{i+c})$ are restricted to the sub-vocabulary $\mathcal{V}_k$ and the restricted functions are denoted as  ${p}_k(x)$ and ${q}_k(x)$, respectively. All function values of ${p}_k(x)$ and ${q}_k(x)$ are loaded from HBM to SRAM.


\circled{\textbf{2}}All necessary partial results are computed with respect to the sub-vocabulary $\mathcal{V}_k$ and stored in SRAM. 
Note that 
the probability ratio $\tau_c(x)$ in~\autoref{eq:rejection}, 
the difference $f(x)=p(x|x_{<i+c})-q(x|x_{<i+c})$ in~\autoref{eq:modified}, and  the numerator $a(x)=\max(0, f(x))$ in~\autoref{eq:max-norm} can be computed in element-wise fashion. 
Their partial evaluations with the sub-vocabulary $\mathcal{V}_k$, denoted as $\tau_{c_k}(x)$, $f_k(x)$, and $a_k(x)$, respectively, do not require any synchronization between threads and blocks, thus allowing fast parallel computation. 
The denominator in~\autoref{eq:max-norm}, $b=\sum_{x^\prime\in\mathcal{V}}\max(0,f(x^\prime)$, requires a reduction across all elements in the vocabulary $\mathcal{V}$ and is more challenging to fully compute in parallel, due to its dependency on other thread blocks.
We use parallel reduction \cite{harris2007parallel_reduce} to compute the partial sum $b_k=\sum_{x^\prime\in\mathcal{V}_k}\max(0, f(x^\prime))$ of the denominator with the sub-vocabulary $\mathcal{V}_k$ in SRAM, and perform the final aggregation across blocks in the subsequent procedure on HBM.

\circled{\textbf{3}}The partial results, $\tau_{c_k}$, $a_k(x)$, and $b_k$, are written back to HBM. 
The partial sum $b_k$ is now combined with the partial sums from other thread blocks to compute the full sum $b$. 
The final division operation to compute $\texttt{max\_norm}(f(x))$ in~\autoref{eq:max-norm} and the resampling procedure in~\autoref{eq:modified} are done once all the partial results are aggregated.

By reorganizing the computations as illustrated in \autoref{fig:approach}, batches of $p({x}| {x}_{<i+c})$ and $q({x}| {x}_{<i+c})$ are loaded only once from HBM into SRAM. 
Moreover, most operations are coupled within the kernel and performed in parallel while using fast SRAM to store intermediate values.  
Only the results necessary to produce token acceptance decisions are written to HBM. 
\subsubsection{Approximated optimization}
\begin{figure*}[ht]
  \centering
\includegraphics[width=1.0\linewidth]{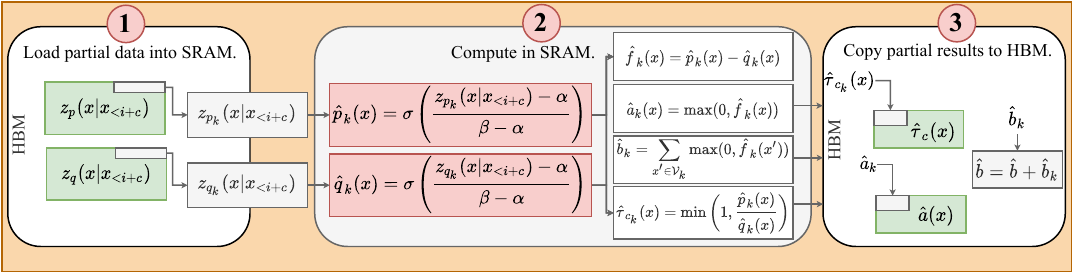}
  \vspace{-0.3in}
  \caption{Overview of the computations within each thread block for sigmoid approximation. Each set of logits is scaled by a minimum constant $\alpha$ and a maximum constant $\beta$. 
 Sigmoid activations $\sigma$ are then computed and stored in SRAM for each segment of draft and target logits. 
 Subsequently, the intermediate values $\hat{f}_k(x)$, $\hat{a}_k(x)$, $\hat{b}_k$, and $\hat{\tau}_{c_k}(x)$ are computed analogous to \autoref{fig:approach}. 
  The resulting outputs are then used to update $\hat{\tau}_c(x)$, $\hat{a}(x)$, and $\hat{b}$ in HBM.}
  \label{fig:sigmoid_sampling}
  \vspace{-0.15in}
\end{figure*}
To further accelerate speculative sampling, we use sigmoid to approximate $p(x| x_{<i+c})$ and $q(x|x_{i+c})$, which are parameterized by softmax in the baseline implementation and the exact method described in~\Cref{ssec:exact}. 
Instead of treating $p(x| x_{<i+c})$ and $q(x|x_{i+c})$ as precomputed inputs to the kernel, the sigmoid approximation is tightly coupled with the other operations in the speculative sampling process. This integration within the kernel substantially accelerates the overall sampling procedure.

\paragraph{Bottleneck of softmax.} For any given input vector $\mathbf{w}=(w_1, \ldots, w_{\lvert\mathcal{V}\rvert})$, the outputs must be positive and they must sum to unity to be interpretable as a probability distribution \cite{bridle89softmax}. 
In softmax, both conditions are satisfied via a normalized exponential transformation. 
With limited value ranges that can be represented in hardware, softmax is prone to overflow or underflow due to the exponentiation. 
Therefore, a numerically stable version is often used~\cite{milakov2018online}: 
\begin{equation}\label{eq:softmax}
\texttt{softmax}(\mathbf{w})_j=\frac{\exp\left(w_j-w_{\texttt{max}}\right)}{\sum_{l=1}^{\lvert\mathcal{V}\rvert} \exp(w_l-w_\texttt{max})} 
\end{equation}
for $j=1, \ldots, \lvert\mathcal{V}\rvert$, where $w_\texttt{max}=\max  \{w_l: 1\leq l \leq \lvert\mathcal{V}\rvert\}$.~\autoref{eq:softmax} requires summing over the size of the vocabulary and finding $w_\texttt{max}$, which makes parallelization on GPUs challenging, since both the summation and $w_\texttt{max}$ require keeping track of intermediate values across blocks~\cite{dao2022flashattention, rabe2022selfattention, wortsman2023replacing}.

The attention mechanism including its softmax computation has been optimized in FlashAttention~\citep{dao2022flashattention} by fusing its operations and using an online algorithm~\citep{milakov2018online,rabe2022selfattention} that splits the workload into blocks and rescales the output of each block. Unlike FlashAttention, we explore a fully local operation that can run without expensive tracking of intermediate variables across blocks, thus allowing for non-blocking parallel computation.  

\paragraph{Sigmoid approximation.} 
Let ${z}_p(x|x_{<i+c})$ be the logits of the target model $M_p^{\texttt{target}}$ given the prefix $(x_1, \ldots, x_{i+c-1})$. 
Similarly, let ${z}_q(x| x_{<i+c})$ be the logits of the draft model $M_q^{\texttt{draft}}$. 
First, we rescale the logits using predefined constant values $\alpha<0$ and $\beta>0$, and then apply sigmoid to these scaled logits to approximate $p(x|x_{i+c})$ and $q(x|x_{<i+c})$ as follows:
\begin{align}
\begin{split}
    \hat{p}(x|x_{<i+c})&=\sigma\left(\frac{z_p(x|x_{<i+c})-\alpha}{\beta-\alpha} \right) \\
    \hat{q}(x|x_{<i+c})&=\sigma\left(\frac{z_q(x|x_{<i+c})-\alpha}{\beta-\alpha} \right),
\end{split}
\label{eq:sigmoid}
\end{align}
where $\sigma(x)=1/(1+\exp(-x))$. 
Although all values of $\hat{p}(x|x_{<i+c})$ and $\hat{q}(x|x_{<i+c})$ are positive, they do not sum to 1 and thus do not constitute valid probability distributions. 
Nonetheless, we rely on these approximations with the hope that they are sufficiently accurate for guiding token acceptance decisions.
Using the approximations $\hat{p}$ and $\hat{q}$, we accept the draft token $x_{i+c}$ sampled from $M_q^{\texttt{draft}}$ if $r_{c}\leq \hat{\tau}_{c}(x_{i+c})$:
\begin{equation*}
\begin{gathered}
    \hat{\tau}_{c}(x_{i+c}) = \min\left(1, \frac{\hat{p}(x_{i+c}| x_{<i+c})}{\hat{q}(x_{i+c}| x_{<i+c})} \right) \\
    r_c\sim U(0,1),
\end{gathered}
\end{equation*}
which is the approximation of~\autoref{eq:rejection}. 
If the token $x_{i+c}$ is rejected, we resample a token from a distribution that approximates~\autoref{eq:modified}:
\begin{equation*}
\begin{gathered}
{x}_{i+c} \sim \texttt{max\_norm}\left(\hat{p}({x}|{x}_{<i+c}) - \hat{q}({x}|{x}_{<i+c})\right). 
\end{gathered} 
\end{equation*}

~\autoref{fig:sigmoid_sampling} illustrates the computations with sigmoid approximation executed in parallel within each thread block. 
The main changes are highlighted in red rectangles. 
\circledred{\textbf{1}}Let ${z}_{p_k}(x| x_{<i+c})$ and ${z}_{q_k}(x| x_{<i+c})$ be restrictions of the logits ${z}_p(x| x_{<i+c})$ and ${z}_q(x| x_{<i+c})$ to the sub-vocabulary $\mathcal{V}_k$ of the corresponding current tile. 
The function values of ${z}_{p_k}(x| x_{<i+c})$ and ${z}_{q_k}(x| x_{<i+c})$ evaluated on $\mathcal{V}_k$ are loaded from HBM into SRAM. 
\circledred{\textbf{2}}We apply sigmoid to the rescaled logits ${z}_{p_k}(x| x_{<i+c})$ and ${z}_{q_k}(x| x_{<i+c})$. 
Since the computation of sigmoid is an element-wise operation and does not depend on values from other threads and blocks, we can execute it in parallel, thereby further accelerating speculative sampling. Similar to step \circled{\textbf{2}} of the exact optimization in~\autoref{fig:approach}, the partial results, $\hat{f}_k(x)$, $\hat{a}_k(x)$, $\hat{b}_k$, and $\hat{\tau}_{c_k}(x)$, which are approximations of $f_k(x)$, $a_k(x)$, $b_k$, and $\tau_{c_k}(x)$, respectively, are computed and stored in SRAM. \circledred{\textbf{3}}
The partial results are written back to HBM, and for $\hat{b}_k$, they are aggregated across blocks to compute the final result $\hat{b}$, which is approximation of $b$.

%% file: 4_exp.tex
\section{Experiments}\label{sec:exp}
\subsection{Experimental setup}

\begin{table*}[t]
\setlength{\tabcolsep}{6.5pt}
\vspace{-0.15in}
\small 
\scalebox{1.0}{
\begin{tabular}{lc|cc|ccc|cc}
\toprule
\multirow{2}{*}{\textbf{Dataset}} & \multirow{2}{*}{\textbf{Subset}} & \multicolumn{2}{c|}{\textbf{Model}} & \multicolumn{3}{c|}{\textbf{WER $(\downarrow)$}} & \multicolumn{2}{c}{$\Delta$\% \textbf{Profiling Time}} \\
\cmidrule(lr){3-4} \cmidrule(lr){5-7} \cmidrule(lr){8-9}
  &  & Target & Draft & Baseline & Exact & Sigmoid & Exact & Sigmoid \\
\midrule
\multirow{2}{*}{LibriSpeech} & clean & \multirow{4}{*}{\makecell{Whisper \\ Small.EN}} & \multirow{4}{*}{\makecell{Distil- \\ Whisper \\ Small.EN}} & 0.08 & 0.08 & 0.09 & 11.7\% & 71.9\% \\
 & other &  & & 0.14 & 0.14 & 0.15 & 10.7\% & 74.6\% \\
TED-LIUM & release3 &  & & 0.22 & 0.22 & 0.24 & 12.5\% & 78.9\% \\
CV16 & en &  &  & 0.22 & 0.22 & 0.27 & 10.4\% & 66.8\% \\
\hline
\multirow{2}{*}{LibriSpeech}  & clean & \multirow{4}{*}{\makecell{Whisper \\ Large V2 }} & \multirow{4}{*}{\makecell{Distil- \\ Whisper \\ Large V2}} & 0.07 & 0.07 & 0.15 & 10.8\% & 83.9\% \\
 & other &  &  & 0.12 & 0.12 & 0.19 & 11.4\% & 84.0\% \\
TED-LIUM   & release3 &  & & 0.20 & 0.20 & 0.23 & 11.4\% & 83.3\% \\
CV16 & en  & &  & 0.25 & 0.25 & 0.31 & \phantom{0}8.7\% & 78.5\% \\
\bottomrule
\toprule
\multirow{2}{*}{\textbf{Dataset}} &  \multirow{2}{*}{\textbf{Subset}} & \multicolumn{2}{c|}{\textbf{Model}} & \multicolumn{3}{c|}{\textbf{ROUGE-1 $(\uparrow)$}} & \multicolumn{2}{c}{$\Delta$\% \textbf{Profiling Time}} \\
\cmidrule(lr){3-4} \cmidrule(lr){5-7} \cmidrule(lr){8-9}
& & Target & Draft & Baseline & Exact & Sigmoid & Exact & Sigmoid \\
\midrule
\multirow{4}{*}{CNN/DM} & -- & Llama2 7B & Sheared Llama 1.3B & 0.30 & 0.30 & 0.26 & \phantom{0}5.7\% & 78.0\% \\
 & -- & Llama2 13B & Sheared Llama 1.3B & 0.31 & 0.31 & 0.29 & 10.1\% & 37.2\% \\
 & -- & Qwen 7B & Qwen 0.5B & 0.31 & 0.31 & 0.26 & \phantom{0}6.8\% & 92.4\% \\
 & -- & Gemma 7B & Gemma 2B & 0.23 & 0.23 & 0.17 & 10.6\% & 68.4\% \\
\hline
\multirow{4}{*}{Xsum} & -- & Llama2 7B & Sheared Llama 1.3B & 0.20 & 0.20 & 0.18 & 11.1\% & 83.6\% \\
    & -- & Llama2 13B & Sheared Llama 1.3B & 0.20 & 0.20 & 0.17 & 10.5\% & 45.7\% \\
 & -- & Qwen 7B & Qwen 0.5B & 0.18 & 0.18 & 0.13 & \phantom{0}7.8\% & 93.6\% \\
 & -- & Gemma 7B & Gemma 2B & 0.18 & 0.18 & 0.13 & \phantom{0}9.9\% & 63.5\% \\
\bottomrule
\end{tabular}
}
\vspace{-0.1in}
\caption{Accuracy and profiling results on ASR and text summarization. 
The column ``$\Delta$\% Profiling Time'' measures the relative reduction in GPU time achieved with our optimized approaches (\textit{exact} and \textit{sigmoid approximation}) compared to the baseline.}
\label{tab:asr_combined}
\vspace{-0.1in}
\end{table*}
\paragraph{Datasets and metrics.}
We evaluate  accuracy and inference speed of our optimized speculative sampling on ASR and single-document summarization. 
For ASR, we measure word error rates (WERs) on the test portions of three English datasets: CommonVoice 16 (CV16)~\cite{ardila2020cv}, LibriSpeech~\cite{panayotov2015librispeech}, and TED-LIUM~\cite{rousseau2012ted}. 
For summarization, we use the test portions of Extreme Summarization (Xsum) \cite{narayan2018xsum} and CNN/Daily Mail (CNN/DM) \cite{nallapati2016cnn} to evaluate the quality of summaries generated by language models with ROUGE-1~\citep{lin-2004-rouge}. 
Additional dataset details are provided in~\Cref{app:dataset_detail}. 

For all tasks, we use the PyTorch~\citep{paszke2019pytorch} profiling tool to obtain execution times for performing speculative sampling.  
We measure the execution time within the entire call stack of the speculative sampling function, including any nested function call (\eg~softmax). The profiling times are summed over all decoding steps and examples in a dataset, before the relative improvement is calculated.

\paragraph{Hyperparameters.}
We set the batch size $B$ to 1 and employ the same heuristic used in the baseline speculative sampling implementation in the Transformers library~\citep{wolf2020transformers}, to determine the number of draft tokens $\gamma$. 
Initially, $\gamma$ is set to 5 and increases by 2 if all speculative tokens sampled from the draft model are accepted; otherwise, it decreases by 1.
We set the maximum sequence length to 256 tokens for ASR and 100 tokens for summarization. 
For ASR, using sigmoid approximation, $\alpha$ and $\beta$ are set to $-10^3$ and $10^3$, respectively. 
In summarization experiments, we use $\alpha=-10^4$ and $\beta=10^4$.  
We set $n = 1024$, \ie, the maximum available threads per block on the NVIDIA A100 GPU. 
\paragraph{Target models.}
We employ Whisper~\citep{radford2022whisper} as the target model series for the ASR task. 
We use both the multilingual 1.55B parameter \texttt{whisper-large-v2} version and the English-only 244M parameter \texttt{whisper-small.en} version of the model. 
For the summarization task, we use Llama2 7B/13B~\citep{touvron2023llama2}, Qwen 1.8B/7B~\citep{bai2023qwen}, and Gemma 7B~\citep{gemmateam2024gemma}. 
More details on the target models are provided in~\Cref{app:tgt_models}. 

\paragraph{Draft models.} Following~\citet{leviathan2023decoding}, \citet{chen2023sampling}, and \citet{zhou2024distillspec} we either use smaller models of the same series or distilled versions of the target model for drafting. 
The draft model family for the ASR task is Distil-Whisper~\citep{gandhi2023distilwhisper}. 
In particular, we use the 166M parameter \texttt{small.en} and the 756M parameter \texttt{distil-large-v2} versions.  
The draft model for Llama2 is Sheared-LLaMA~\citep{xia2024sheared}, a version of Llama2 pruned to 1.3B parameters. 
The draft models corresponding to Qwen and Gemma are the 500M and 2B parameter versions of the same series. 
More details on the draft models are provided in~\Cref{app:draft_models}. 

\paragraph{Implementation details.} 
We use the implementation of speculative sampling provided by the Transformers library~\citep{wolf2020transformers} (v4.38.2) in conjunction with PyTorch (v2.2.2) as our baseline. 
Unless stated otherwise, all models are loaded in FP16 and executed on A100 GPUs with 80GB HBM using the same compute node running CUDA 12.3 and NVIDIA device driver version 545. 

\begin{figure*}[t]
    \centering
    \begin{subfigure}[b]{0.85\textwidth}
        \centering
        \includegraphics[width=\textwidth]{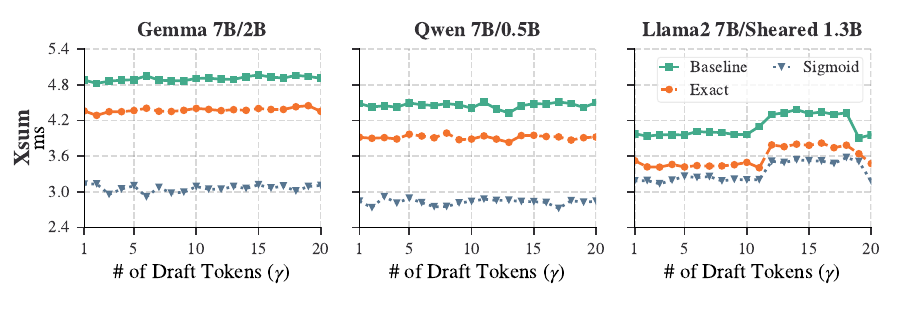}
        \vspace{-0.4in}
        \caption{}
        \label{fig:mean_sampling_time-text}
        \vspace{-1mm}
    \end{subfigure}
    \vspace{-5mm}
    \begin{subfigure}[b]{0.82\textwidth}
        \centering
\includegraphics[width=\textwidth]{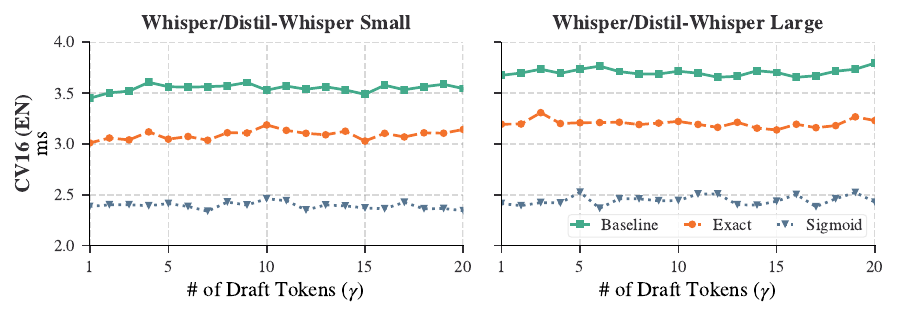}
        \vspace{-0.35in}
        \caption{}
        \label{fig:asr_mean_sampling_time-asr}
    \end{subfigure}
    \captionsetup{width=0.8\textwidth}
    \vspace{0.05in}
    \caption{Average execution time of the speculative sampling algorithm per decoding step for varying initial $\gamma$ values on randomly sampled subsets (10\%) of Xsum and CV16 test sets.}
    \label{fig:mean_sampling_time}
    \vspace{-4mm}
\end{figure*}

\subsection{Main results}
\autoref{tab:asr_combined} summarizes accuracy metrics and profiling results for the ASR and  text summarization tasks. 
The table details the datasets, target and draft models used, performance metrics, and the relative reduction in overall GPU profiling time achieved by our optimized approaches (exact and sigmoid approximation) compared to the baseline.

In the ASR task, our exact optimization method maintains the same WER compared to the baseline and achieves reduction in profiling time ranging from 8.7\% to 12.5\%. 
The sigmoid approximation approach results in moderately increased WERs, but yields more significant profiling time improvements, ranging from 71.9\% to 84.0\%.

In the text summarization task, our exact optimization method demonstrates a similar trend as observed in the previous ASR experiments, reducing profiling time by 5.7\% to 11.1\% without affecting ROUGE-1 scores. 
The non-exact sigmoid approximation further achieves significant profiling time reductions, reaching up to 93.6\%.
However, we also observe an absolute difference in ROUGE-1 of 0.02 to 0.06 points.

Additionally, we provide relative wall-clock time improvements for the overall text generation process in~\autoref{tab:walltime} of \Cref{app:mean_time}, showing that the results obtained via profiling translate into improvements in wall-clock time for both the exact and the sigmoid approximation method.

\subsection{Analysis and discussion}
\paragraph{Execution times remain stable over varying $\gamma$.}
To assess the robustness of our exact and sigmoid optimization methods, we measure execution times across different models and varying numbers of initial draft tokens $\gamma$. 
For text summarization, we randomly sample 10\% of the Xsum test set and use Gemma, Qwen, and Llama2 model combinations to generate summaries. 
For ASR, we use 10\% of randomly sampled examples from the CV16 test set. 
~\autoref{fig:mean_sampling_time-text} and ~\autoref{fig:asr_mean_sampling_time-asr} illustrate the average execution times of the different implementations profiled per decoding step. 
Both figures show average execution times measured in milliseconds (ms) for the number of draft tokens ranging from 1 to 20. 
The average execution times for the optimized approaches (exact and sigmoid) are consistently below the baseline across all models and values of $\gamma$. 
Furthermore, the execution times of the optimized approaches are stable across different choices of $\gamma$ for the Gemma and Qwen models, whereas the Llama2 7B/Sheared LLaMA 1.3B model combination exhibits small sensitivity to the number of draft tokens. 

As depicted in~\autoref{fig:asr_mean_sampling_time-asr}, the ASR models also exhibit stable execution times across different $\gamma$, further validating the robustness of our optimization methods with varying numbers of draft tokens.

\begin{figure}[t]
\centering
\includegraphics[width=1.0\linewidth]{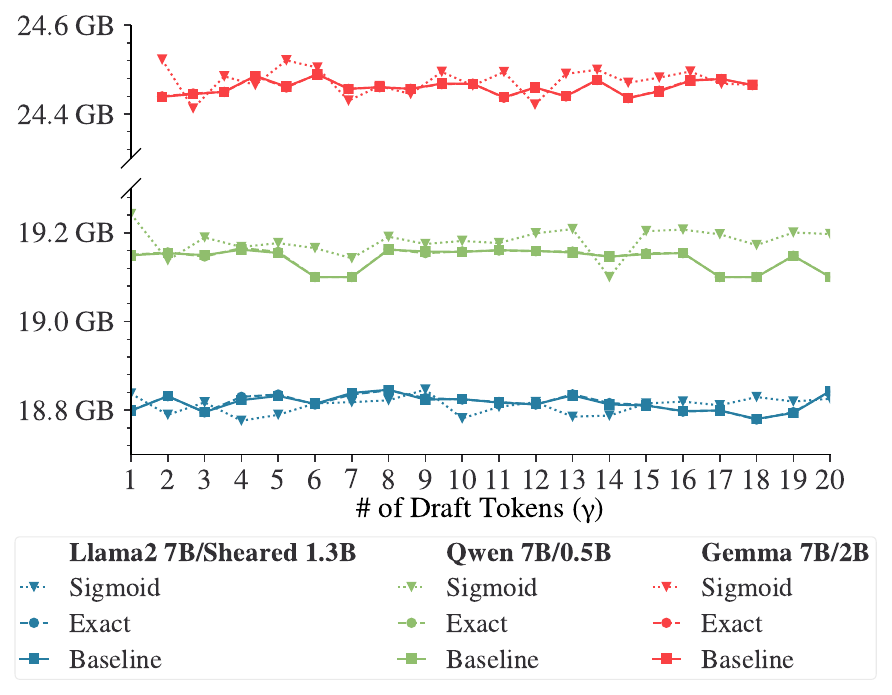}
\vspace{-0.3in}
  \caption{Peak memory usage (HBM) on randomly sampled 10\% of the \textbf{Xsum} test set for varying initial values of $\gamma$. 
  }
\label{fig:peak_memory}
\vspace{-0.2in}
\end{figure}

\begin{figure}[t]
\centering
\includegraphics[width=1.0\linewidth]{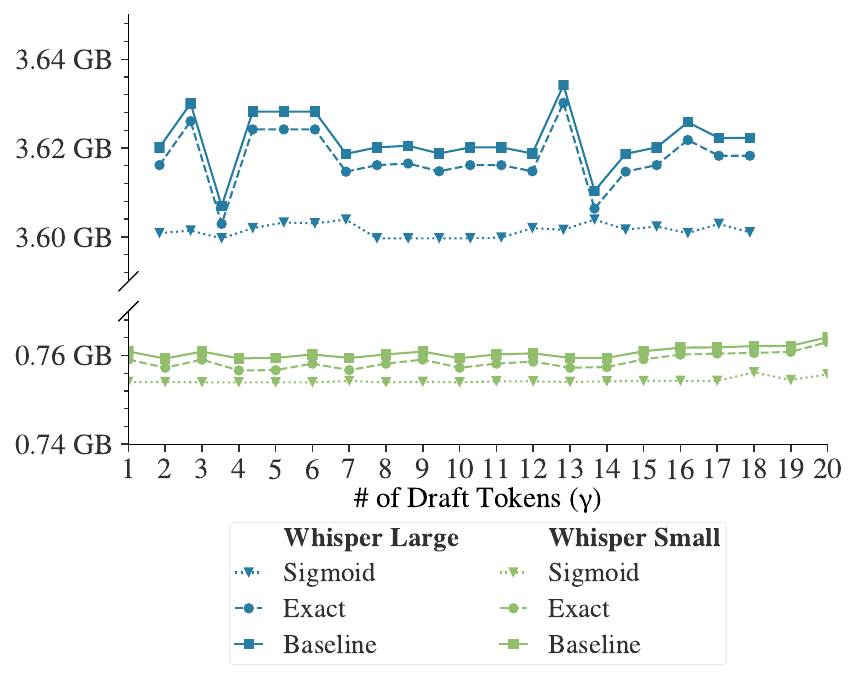}
\vspace{-0.35in}
\caption{Peak memory usage (HBM) on randomly sampled 10\% of the \textbf{CV16} test set for varying initial values of $\gamma$.}
\label{fig:peak_memory_asr}
\vspace{-0.03in}
\end{figure}

\paragraph{Optimized sampling does not introduce additional memory overhead.}
We assess the memory usage of our optimized methods relative to the baseline implementation. 
\autoref{fig:peak_memory} shows the peak memory usage (HBM) on randomly sampled instances  (10\%) of the Xsum test set with various initial values of $\gamma$ and different language models.
The graph indicates that our optimized approaches do not introduce additional memory overhead compared to the baseline. 
For all three model combinations (Llama2, Qwen, and Gemma), the memory usage of the optimized methods fluctuates slightly (within a range of approximately 200MB) around the memory usage of the baseline across all draft tokens. 
As illustrated in~\autoref{fig:peak_memory_asr}, the memory usage results for the Whisper models display a pattern consistent with the text summarization experiments, showing fluctuation within a range of under 10 MB.

\paragraph{Effect of scaling logits.}
To study the effect of logit scaling in the sigmoid approximation method, we compare profiling time and performance metrics under varying values of $\alpha$ and $\beta$ (cf.~\autoref{eq:sigmoid}). 
\autoref{tab:scale_ablation} provides a comparison of different scaling factors for  subsets (10\%) of CV16 and Xsum using Whisper Small.EN and Llama2 7B, respectively.   
For each task, the table shows the values of $\alpha$ and $\beta$ applied, the resulting WER or ROUGE-1 score, and the relative improvement in profiling time over the non-optimized baseline implementation.  
Note that scaling is necessary due to the numerical instability induced by the exponentiation in the sigmoid function. 

The Llama2 model combination exhibits relative stability across different scaling factors, showing minor fluctuations in ROUGE-1 scores and profiling time improvements, whereas the Whisper model combination is more sensitive, with scaling factors of $\pm10^5$ leading to substantial deterioration of both WER and profiling time. 
This can be attributed to the logits of Whisper models being generated in half precision, whereas the logits generated by the Llama models are available in full precision. 
However, we also find that scaling factors of $\pm10^3$ and $\pm10^4$ generally yield comparable results in both accuracy and profiling time improvement across the model combinations investigated in this work. 
The results of the same analysis for the other draft and target model combinations are provided in~\autoref{tab:scale_ablation_full} from ~\Cref{app:scale_ablation}. 

Furthermore, we see a general trend where higher accuracy (lower WER and higher ROUGE-1) coincides with higher profiling time improvements.
 This relationship is due to the token acceptance process, where better calibrated models accept more tokens, requiring fewer executions of resampling and fewer overall calls of the speculative sampling kernel. 

\begin{table}[t]
\centering
\setlength{\tabcolsep}{3pt}
\resizebox{0.49\textwidth}{!}{
\begin{tabular}{c|cc|c|c}
\toprule
\textbf{Target/Draft}  & \multicolumn{2}{c|}{\textbf{Scale} ($\alpha, \beta$)}  & \textbf{WER $(\downarrow)$} & $\Delta$\% \textbf{Prof. Time} \\
\midrule
\multirow{5}{*}{\makecell{Whisper Small.EN/ \\ Distil-Whisper \\ Small.EN}} & \multicolumn{2}{c|}{Baseline} & 0.24 & -- \\
 & $-10^1$ & $10^1$ & 0.41 & 24.0\% \\
 & $-10^3$ & $10^3$ & 0.28 & 59.5\% \\
 & $-10^4$ & $10^4$ & 0.26 & 64.6\% \\
 & $-10^5$ & $10^5$ & 29.34 & -10826.1\% \\
\bottomrule
\toprule
\textbf{Target/Draft} &   \multicolumn{2}{c|}{\textbf{Scale} ($\alpha, \beta$)} & \textbf{{ROUGE-1} $(\uparrow)$} & $\Delta$\% \textbf{Prof. Time} \\
\midrule
\multirow{5}{*}{\makecell{Llama2 7B/\\Sheared \\ Llama 1.3B}} & \multicolumn{2}{c|}{Baseline} & 0.19 & -- \\
 & $-10^1$& $10^1$ & 0.16 & 49.8\% \\
 & $-10^3$ & $10^3$ & 0.16 & 53.4\% \\
 & $-10^4$ & $10^4$ & 0.15 & 50.0\% \\
 & $-10^5$ & $10^5$ & 0.16 & 51.9\% \\
\bottomrule
\end{tabular}
}
\vspace{-0.1in}
\caption{Impact of varying $\alpha$ and $\beta$ on accuracy and profiling time of sigmoid approximation on \textbf{CV16} and \textbf{Xsum}. 
}
\label{tab:scale_ablation}
\vspace{-0.15in}
\end{table}

\paragraph{Data transfer between HBM and SRAM.} 
To assess efficiency in terms of data movement between HBM and on-chip SRAM, we compare the realized bandwidths of each implementation. 
The bandwidth is calculated by dividing the total bytes transferred by the execution time. 
The number of bytes transferred is derived from the total number of sectors moved between HBM and SRAM, with each sector consisting of 32 bytes. Execution time refers to the duration during which the GPU is actively running a kernel, i.e., when at least one GPU unit is engaged in computation rather than idling, waiting, or stalled.
A lower realized bandwidth indicates a reduced communication overhead between HBM and SRAM. 

The results in \autoref{tab:data_transfer} show lower memory bandwidth usage for the Qwen and Gemma models with the exact optimization approach compared to the baseline implementation. 
However, in the case of the Llama2 and Whisper models, despite overall lower realized bandwidths relative to the Qwen and Gemma models, higher bandwidths are observed with the exact optimization compared to their corresponding baseline. 

The sigmoid approximation has consistently higher realized bandwidths across all model combinations compared to the baseline. 
Although the sigmoid optimization approach reduces the overall amount of data transferred, \ie, the total number of bytes moved between HBM and SRAM, due to the element-wise approximation of the softmax function within the sampling kernel, the significantly faster execution times result in higher overall realized bandwidths. 
However, even the highest realized bandwidths are far below the theoretical HBM bandwidth limit of $\sim$2 TB/s \cite{nvidia_a100_2020}, indicating that memory transfer is not the limiting factor for performance. 
\begin{table}[h!]
\centering
\setlength{\tabcolsep}{3pt}
\resizebox{0.49\textwidth}{!}{
\begin{tabular}{cc|ccc}
\toprule
\multicolumn{2}{c|}{\textbf{Model}} & \multicolumn{3}{c}{\textbf{Realized Bandwidth}} \\ 
\cmidrule(lr){1-2} \cmidrule(lr){3-5}
Draft & Target & Baseline & Exact & Sigmoid \\ 
\midrule
\makecell{Whisper\\Small.EN} & \makecell{Distil-Whisper \\  Small.EN} & 13.98 GB/s & 14.16 GB/s & 16.33 GB/s \\
\makecell{Whisper\\Large V2} & \makecell{Distil-Whisper\\Large V2} & 9.32 GB/s & 11.18 GB/s &  16.06 GB/s\\
\hline
Qwen 7B & Qwen 0.5B & 44.99 GB/s & 31.65 GB/s & 52.14 GB/s \\
Gemma 7B  & Gemma 2B & 53.69 GB/s & 38.51 GB/s & 62.99 GB/s \\
Llama2 7B & Sheared Llama 1.3B & 24.56 GB/s & 27.70 GB/s & 30.52 GB/s \\
Llama2 13B & Sheared Llama 1.3B & 20.18 GB/s & 24.08 GB/s & 31.39 GB/s \\
\bottomrule
\end{tabular}
}
\vspace{-0.1in}
\caption{Comparison of realized bandwidths across models and optimization techniques using 100 examples of the XSum test set for Qwen, Gemma, and Llama2 models and 100 examples of the CV16 test set for Whisper models.}
\label{tab:data_transfer}
\end{table}
\paragraph{Results on RTX 2080 TI GPU.}\label{app:rtx_results}
\autoref{tab:rtx_performance} shows an accuracy and profiling time comparison using RTX 2080 TI GPUs. 
We observe that our optimization methods achieve performance similar to A100 GPUs with relative improvements in profiling time ranging between $\sim$5\% and $\sim$13\% with the exact method and between $\sim$62\% and $\sim$82\% with sigmoid approximation. 
The summarization experiment with Qwen uses a 1.8B/0.5B parameter model combination instead of the 7B/0.5B model combination from the main experiments in \autoref{tab:asr_combined}, due to the limited HBM size of 11GB available on the RTX 2080 TI series. 
\begin{table}[h!]
\setlength{\tabcolsep}{1pt}
\resizebox{0.49\textwidth}{!}{
\begin{tabular}{c|c|c|ccc|ccc}
\toprule
\multirow{2}{*}{\textbf{Dataset}} & \multirow{2}{*}{\textbf{Subset}} & \multirow{2}{*}{\textbf{Target/Draft}} & \multicolumn{3}{c|}{\textbf{WER $(\downarrow)$}} & \multicolumn{2}{c}{$\Delta$ \textbf{Prof. Time}} \\
& & & Basel. & Exact & Sigm. & Exact & Sigmoid \\
\midrule
LibriSpeech & clean   & \multirow{4}{*}{\makecell{Whisper \\ Small.EN/ \\ Distil \\ Small.EN}} & 0.08  & 0.08  & 0.09  & 7.6\%  & 66.8\%  \\
LibriSpeech & other   &    & 0.14  & 0.14  & 0.15  & 12.3\% & 65.9\%  \\
TED-LIUM    & release3 &   & 0.22  & 0.22  & 0.24  & 12.6\% & 62.8\%  \\
CV16        & en      &    & 0.22  & 0.22  & 0.27  & 12.3\% & 62.0\%  \\
\hline
LibriSpeech & clean   & \multirow{4}{*}{\makecell{Whisper \\ Large V2/ \\ Distil \\ Large V2}} & 0.07  & 0.07  & 0.15  & 11.2\% & 82.5\%  \\
LibriSpeech & other   &   & 0.12  & 0.12  & 0.19  & 5.7\%  & 79.2\%  \\
TED-LIUM    & release3 &   & 0.20  & 0.20  & 0.23  & 9.4\%  & 79.8\%  \\
CV16        & en      &    & 0.25  & 0.25  & 0.31  & 7.8\%  & 75.6\%  \\
\midrule
\multirow{2}{*}{\textbf{Dataset}} & \multirow{2}{*}{\textbf{Subset}} & \multirow{2}{*}{\textbf{Target/Draft}} & \multicolumn{3}{c|}{\textbf{ROUGE-1 $(\uparrow)$}} & \multicolumn{2}{c}{$\Delta$ \textbf{Prof. Time}} \\
& &  & Base. & Exact & Sigm. & Exact & Sigmoid \\
\midrule
Xsum    &  --  & \makecell{ Qwen 1.8B \\ Qwen 0.5} & 0.15  & 0.15  & 0.11  & 5.0\%  & 76.8\%  \\
\bottomrule
\end{tabular}
}
\vspace{-0.1in}
\caption{Accuracy and profiling time comparison across various datasets and models using a \textbf{RTX 2080 TI} GPUs.}
\label{tab:rtx_performance}
\end{table}

%% file: 5_conclusion.tex
\section{Conclusions}\label{sec:conclusion}
We introduced two optimization methods to accelerate speculative sampling for autoregressive models on hardware accelerators. 
By computing significant portions of intermediate matrices across multiple GPU threads within thread blocks, our exact optimization method led to improved sampling speed without compromising accuracy. 
Additionally, we employed an approximation technique using element-wise sigmoid instead of softmax, to enable parallel computation of probabilities. 
This approximation further accelerated the decoding process but resulted in a small degradation of sampling quality. 

%% file: 6_limitation.tex
\section*{Limitations}
In this work, we study the inference efficiency of speech and language models in the context of speculative decoding using GPU hardware accelerators. 
Our investigation includes two optimized speculative sampling algorithms, tested on these models to enhance inference speed. 
While GPUs are the most common general-purpose hardware accelerators, there exist purpose-built architectures such as Cerebras's Wafer Scale Engines, Google's TPUs, and GraphCore's IPUs, where the differences in system design may negate or significantly reduce the latency gains. 
Our experiments were conducted exclusively on A100 and RTX 2080 TI GPUs on a single compute node. 
Therefore, the generalizability of the results to other hardware configurations remains uncertain. 
The performance outcomes may be influenced by other hardware and network configurations, such as multi-GPU and multi-node setups, as well as the availability fast interconnects (\eg~Infiniband), and other network conditions. 
Additionally, our study evaluates the effectiveness of the optimized algorithm based on decoding time, and our claims may not translate to other metrics, such as energy usage or heat generation, although they play an important role in real-world production settings.

%% file: 7_appendix.tex
\section{Appendix}\label{sec:appendix}
\subsection{Target model details}\label{app:tgt_models}
\paragraph{Whisper.}
Whisper \cite{radford2022whisper} is a family of models trained to perform multiple tasks such as multilingual ASR, language identification, and speech translation.
The models are trained on $\sim$680k hours of labeled audio data retrieved from the world wide web and are available in five sizes ranging from 39M parameters to 1.55B parameters. 
Utilizing an encoder-decoder Transformer architecture, Whisper receives 80-channel log-Mel spectrogram representations with a 25ms window and a stride of 10ms as inputs. 
We conduct experiments on both the multilingual 1.55B parameter \texttt{whisper-large-v2} version and the English-only 244M parameter \texttt{whisper-small.en} version. 
\paragraph{Llama2.}
Llama2 \cite{touvron2023llama2} is a collection of LLMs ranging from 7B to 70B parameters. 
The models are pretrained on 2 trillion tokens of text data from publicly available sources. 
The architecture is based on Llama1 \cite{touvron2023llama}, utilizing pre-normalization with RMSNorm \cite{zhang2019rmsnorm}, SwiGLU \cite{shazeer2020glu} activation functions, and rotary positional embeddings (RoPE) \cite{su2023roformer}. 
Notable architectural changes include an expanded context length of 4K tokens and the adoption of grouped-query attention (GQA) for the 34B and 70B models. 
We employ the 7B and 13B versions of Llama2 as target models. 
\paragraph{Qwen.}
The Qwen \cite{bai2023qwen} model series offers a range of decoder-only language models with parameter counts between 500M and 110B. 
The models are pretrained on up to 3 trillion tokens of various multilingual text, code, and mathematics resources. 
The architecture is similar to Llama2 with small modifications, such as no weight tying between input embeddings and output projection. 
We employ Qwen v1.5 in our experiments and use the 7B parameter variant as the target model. 
\paragraph{Gemma.}
Gemma \cite{gemmateam2024gemma} comprises two model variants, featuring 2B and 7B parameters, pretrained on 3 trillion and 6 trillion tokens respectively. 
Gemma is based on the Gemini \cite{geminiteam2024gemini} model family. 
The focus is primarily on English text from web documents, mathematics, and code, omitting multimodal capabilities and optimization for multilingual tasks. 
Similar to Llama2, the Gemma models leverage RoPE and RMSNorm, and embeddings are shared across inputs and outputs to reduce model size. 
We use the 7B parameter variant of Gemma v1.0 as the target model. 

\subsection{Draft model details}\label{app:draft_models}
\paragraph{Distil-Whisper.} 
The draft model series for the ASR task is Distil-Whisper \cite{gandhi2023distilwhisper}, a collection of smaller versions of the Whisper model. 
Distil-Whisper applies knowledge distillation \cite{hinton15distill} to emulate the performance of the original Whisper model using a large ($\approx$21k hours) pseudo-labeled training corpus. 
The distilled models aim to maintain the robustness of Whisper towards varying audio domains and noisy acoustic conditions and are designed
to be paired with Whisper in a speculative decoding setting. 
We use the 166M parameter \texttt{small.en} version as the draft model for the small 244M parameter target model and the 756M parameter \texttt{distil-large-v2} version as the draft model for the large 1.55B parameter target model. 
\paragraph{Sheared-LLaMA.}
The draft model series for our experiments with Llama2 is Sheared-LLaMA \cite{xia2024sheared}. 
Sheared-LLaMA utilizes a structured pruning approach to reduce the size of the 7B parameter Llama 2 model to 1.3B and 2.7B parameters. 
The structured pruning approach removes parameters from the source model until a given target configuration is satisfied. 
Learned pruning masks representing discrete prune or retain decisions are used to create a smaller sub-network matching the specified target configuration. 
We employ the 1.3B version of Sheared-LLaMA in our experiments.

\begin{table*}[t]
\small
\setlength{\tabcolsep}{11pt}
\vspace{-0.1in}
\begin{tabular}{cc|cc|cc}
\toprule
\multirow{2}{*}{\textbf{Dataset}} & \multirow{2}{*}{\textbf{Subset}} & \multicolumn{2}{c|}{\textbf{Model}} & \multicolumn{2}{c}{\textbf{Wall-clock Improvement}} \\
 &  & Target & Draft & Exact & Sigmoid \\
\midrule
\multirow{2}{*}{TED-LIUM} & release3 & Whisper Small.EN & Distil-Whisper Small.EN & 8.7\% & 58.7\% \\
 & release3 & Whisper Large V2 & Distil-Whisper Large V2 & 6.6\% & 66.3\% \\
\hline
\multirow{4}{*}{LibriSpeech} & clean & Whisper Small.EN & Distil-Whisper Small.EN & 5.8\% & 49.5\% \\
 & clean & Whisper Large V2 & Distil-Whisper Large V2 & 7.4\% & 65.3\% \\
 & other & Whisper Small.EN & Distil-Whisper Small.EN & 7.1\% & 54.5\% \\
 & other & Whisper Large V2 & Distil-Whisper Large V2 & 7.0\% & 66.2\% \\
\hline
\multirow{2}{*}{CV16} & en & Whisper Small.EN & Distil-Whisper Small.EN & 8.1\% & 51.8\% \\
 & en & Whisper Large V2 & Distil-Whisper Large V2 & 5.4\% & 61.3\% \\
\hline\hline
\multirow{4}{*}{CNN/DailyMail}  & -- & Gemma 7B & Gemma 2B & 3.0\% & 24.2\% \\
 & -- & Qwen 7B & Qwen 0.5B & 1.6\% & 39.8\% \\
 & -- & Llama2 7B & Sheared Llama 1.3B & 4.2\% & 18.2\% \\
 & -- & Llama2 13B & Sheared Llama 1.3B & 2.6\% & 23.5\% \\
\hline
\multirow{4}{*}{Xsum} & -- & Gemma 7B & Gemma 2B & 1.2\% & 18.5\% \\
 & -- & Qwen 7B & Qwen 0.5B & 4.3\% & 59.1\% \\
 & -- & Llama2 7B & Sheared Llama 1.3B & 6.5\% & 53.1\% \\
 & -- & Llama2 13B & Sheared Llama 1.3B & 10.9\% & 23.6\% \\
\bottomrule
\end{tabular}
\vspace{-0.1in}
\caption{Relative wall-clock time improvements for both exact and sigmoid sampling on all tasks and model combinations. Wall-clock time measures the total time spent in the speculative decoding loop, including all forward passes through the draft and target models. The relative improvements are computed based on the total time required to perform speculative decoding for the full dataset.  }
\label{tab:walltime}
\end{table*}
\subsection{Dataset details}\label{app:dataset_detail}
\paragraph{ASR.}
We used the test sets of three English ASR benchmark datasets: CommonVoice v16 \cite{ardila2020cv}, LibriSpeech \cite{panayotov2015librispeech}, and TED-LIUM \cite{rousseau2012ted} for the ASR task. 
The data comprises multiple domains such as audiobooks, political speeches, interviews, and narrated Wikipedia articles.
The utterance lengths vary between 0.2 and 330 seconds with an average duration of 7.6$\pm$6.6 seconds. 

\paragraph{Text summarization.}
We used two datasets for text summarization: Extreme Summarization (Xsum) \cite{narayan2018xsum} and CNN/Daily Mail (CNN/DM) \cite{nallapati2016cnn}. 
The Xsum test set contains 11,334 online articles from the British Broadcasting Corporation (BBC) and the CNN/DM test set contains 11,490 news articles published by CNN and the Daily Mail. 
We performed 0-shot evaluation for CNN/DM and Xsum, and used the ROUGE-1 metric for comparison. 
To prompt the model for a summary, we placed ``\texttt{Summary:}'' after each input article. 
Summaries were generated with a maximum token length of 100 for both Xsum and CNN/DM. 

\subsection{Wall-clock time improvement}
\autoref{tab:walltime} summarizes the relative wall-clock time improvements for the overall text generation process. 
Both the exact and the sigmoid approximation method translate into relative improvements compared to the baseline implementation. 
Wall-clock times are less precise, since they also include the forward passes through the draft and target models, which may lead to additional overhead introduced by the deep learning framework \cite{fernandez2023frameworktax}, and the time spent on CPU, which does not take varying rates of context switches and stalling due to execution of higher-priority processes into account.

\subsection{Average times per decoding step}\label{app:mean_time}
The average times spent in the speculative sampling procedure per decoding step are summarized in \autoref{tab:mean_time}. 
Our implementation achieved consistently lower average sampling times than the reference implementation. 
While the the average sampling time was generally longer for the text generation tasks, the average times with our implementation were still consistently lower than the reference implementation.  
\begin{table*}[h!]
\small 
\resizebox{\textwidth}{!}{
\setlength{\tabcolsep}{2pt}
\begin{tabular}{cc|cc|ccc|cc}
\toprule
\multirow{2}{*}{\textbf{Dataset}}  & \multirow{2}{*}{\textbf{Subset}}  & \multicolumn{2}{c|}{\textbf{Model}} & \multicolumn{3}{c|}{\textbf{Avg.}$\pm$\textbf{Std. (ms)}} & \multicolumn{2}{c}{$\Delta$\% \textbf{Prof. Time}} \\
 &  & Target & Draft & Baseline & Exact & Sigmoid & Exact & Sigmoid \\
\midrule
\multirow{2}{*}{TED-LIUM} & release3 & Small.EN & Distil Small.EN & 4.17$\pm$0.81 & 3.67$\pm$0.64 & 3.12$\pm$1.09 & 11.9\% & 25.1\% \\
 & release3 & Large V2 & Distil Large V2 & 4.37$\pm$0.58 & 3.88$\pm$0.46 & 3.62$\pm$1.09 & 11.2\% & 17.2\% \\
\hline
\multirow{4}{*}{LibriSpeech} & clean & Small.EN & Distil Small.EN & 4.31$\pm$1.00 & 3.81$\pm$0.80 & 3.15$\pm$1.13 & 11.7\% & 27.0\% \\
 & clean & Large V2 & Distil Large V2 & 4.35$\pm$0.60 & 3.88$\pm$0.52 & 3.55$\pm$1.08 & 10.8\% & 18.4\% \\
 & other & Small.EN & Distil Small.EN & 4.14$\pm$0.84 & 3.68$\pm$0.68 & 3.20$\pm$1.07 & 11.1\% & 22.7\% \\
 & other & Large V2 & Distil Large V2 & 4.39$\pm$0.61 & 3.90$\pm$0.49 & 3.55$\pm$1.07 & 11.2\% & 19.1\% \\
\hline
\multirow{2}{*}{CV16} & en & Small.EN & Distil Small.EN & 4.14$\pm$0.85 & 3.71$\pm$0.70 & 3.33$\pm$1.10 & 10.4\% & 19.8\% \\
 & en & Large V2 & Distil Large V2 & 4.37$\pm$0.61 & 3.92$\pm$0.51 & 3.62$\pm$1.07 & 10.4\% & 17.2\% \\
\hline\hline
\multirow{4}{*}{CNN/DailyMail} & -- & Gemma 7B & Gemma 2B & 6.54$\pm$0.47 & 5.84$\pm$0.39 & 4.37$\pm$0.54 & 10.6\% & 33.1\% \\
 & -- & Qwen 7B & Qwen 0.5B & 12.01$\pm$1.08 & 11.20$\pm$1.03 & 3.34$\pm$0.45 & 6.8\% & 72.2\% \\
 & -- & Llama2 7B & Sheared Llama 1.3B & 11.33$\pm$0.94 & 10.69$\pm$0.83 & 3.65$\pm$0.62 & 5.7\% & 67.8\% \\
 & -- & Llama2 13B & Sheared Llama 1.3B & 3.99$\pm$0.52 & 3.59$\pm$1.70 & 3.26$\pm$0.26 & 10.1\% & 18.3\% \\
\hline
\multirow{4}{*}{Xsum} & -- & Gemma 7B & Gemma 2B & 6.39$\pm$0.50 & 5.76$\pm$0.42 & 4.61$\pm$0.55 & 9.9\% & 27.9\% \\
 & -- & Qwen 7B & Qwen 0.5B & 11.55$\pm$1.39 & 10.65$\pm$1.51 & 3.20$\pm$0.42 & 7.8\% & 72.3\% \\
 & -- & Llama2 7B & Sheared Llama 1.3B & 4.66$\pm$0.42 & 4.14$\pm$2.88 & 3.64$\pm$0.56 & 11.1\% & 21.8\% \\
 & -- & Llama2 13B & Sheared Llama 1.3B & 4.67$\pm$0.41 & 4.17$\pm$1.70 & 3.70$\pm$0.44 & 10.5\% & 20.7\% \\
\bottomrule
\end{tabular}
}
\vspace{-0.1in}
\caption{Average time and standard deviation spent within the speculative sampling algorithm \textbf{per decoding step}. 
The column ``$\Delta$\% Prof. Time'' measures the relative reduction in average time per decoding step (``Baseline'' vs. ``Exact'' and``Sigmoid''). Scaling constants for sigmoid approximation: $\alpha=-10^3$ and $\beta=10^3$ for ASR, $\alpha=-10^4$ and $\beta=10^4$ for summarization.
}
\label{tab:mean_time}
\end{table*}

\subsection{Effect of scaling logits}\label{app:scale_ablation}
\autoref{tab:scale_ablation_full} shows the impact of various logit scaling factors on performance and profiling time of sigmoid approximation on CV16 and Xsum. 
The values are computed on a random sample of 10\% of each dataset. 
\begin{table}[h!]
\centering
\vspace{-0.15in}
\setlength{\tabcolsep}{2pt}
\small 
\resizebox{0.5\textwidth}{!}{
\begin{tabular}{c|cc|c|c}
\toprule
\textbf{Draft/Target}  &  \multicolumn{2}{c|}{\makecell{\textbf{Scale} \\ ($\alpha, \beta$)}} & \textbf{WER $(\downarrow)$} & \makecell{\textbf{Prof. Time} \\ $\Delta$\% } \\
\midrule
\multirow{5}{*}{\makecell{Whisper Small.EN \\ Distil Small.EN}} & \multicolumn{2}{c|}{Baseline} & \phantom{0}0.24 & -- \\
 & $-10^1$ & $10^1$ & \phantom{0}0.41 & 24.0\% \\
 & $-10^3$ & $10^3$ & \phantom{0}0.28 & 59.5\% \\
 & $-10^4$ & $10^4$ & \phantom{0}0.26 & 64.6\% \\
 & $-10^5$ & $10^5$ & 29.34 & -10826.1\% \\
\hline
\multirow{5}{*}{\makecell{Whisper Large V2/ \\ Distil Large V2} }
 & \multicolumn{2}{c|}{Baseline} & \phantom{0}0.24 & -- \\
 & $-10^1$ & $10^1$ & \phantom{0}0.42 & 64.3\% \\
 & $-10^3$ & $10^3$ & \phantom{0}0.34 & 75.0\% \\
 & $-10^4$ & $10^4$ & \phantom{0}0.31 & 78.6\% \\
 & $-10^5$ & $10^5$ & 30.91 & -4458.3\% \\
\bottomrule
\toprule
\textbf{Draft/Target} &  \multicolumn{2}{c|}{\makecell{\textbf{Scale} \\ ($\alpha, \beta$)}} & \textbf{\scriptsize{ROUGE-1} $(\uparrow)$} & \makecell{\textbf{Prof. Time} \\ $\Delta$\% } \\
\midrule
\multirow{5}{*}{\makecell{Gemma 7B/ \\Gemma 2B}}
 & \multicolumn{2}{c|}{Baseline} & 0.17 & -- \\
 & $-10^1$ & $10^1$ & 0.01 & -124.6\% \\
 & $-10^3$ & $10^3$ & 0.13 & 66.1\% \\
 & $-10^4$ & $10^4$ & 0.13 & 71.3\% \\
 & $-10^5$ & $10^5$ & 0.14 & 73.0\% \\
\hline
\multirow{5}{*}{\makecell{Qwen 7B/\\Qwen 0.5B}} & \multicolumn{2}{c|}{Baseline} & 0.18 & -- \\
 & $-10^1$ & $10^1$ & 0.09 & 57.6\% \\
 & $-10^3$ & $10^3$ & 0.12 & 71.4\% \\
 & $-10^4$ & $10^4$ & 0.12 & 71.5\% \\
 & $-10^5$ & $10^5$ & 0.11 & 71.7\% \\
\hline
\multirow{5}{*}{\makecell{Llama2 7B/\\Sheared 1.3B}} & \multicolumn{2}{c|}{Baseline} & 0.19 & -- \\
 & $-10^1$ & $10^1$ & 0.16 & 49.8\% \\
 & $-10^3$ & $10^3$ & 0.16 & 53.4\% \\
 & $-10^4$ & $10^4$ & 0.15 & 50.0\% \\
 & $-10^5$ & $10^5$ & 0.16 & 51.9\% \\
\hline
\multirow{5}{*}{\makecell{Llama2 13B/\\Sheared 1.3B}} & \multicolumn{2}{c|}{Baseline} & 0.20 & -- \\
 & $-10^1$ & $10^1$ & 0.15 & 46.9\% \\
 & $-10^3$ & $10^3$ & 0.17 & 46.8\% \\
 & $-10^4$ & $10^4$ & 0.15 & 46.7\% \\
 & $-10^5$ & $10^5$ & 0.16 & 45.9\% \\
\bottomrule
\end{tabular}
}
\vspace{-0.1in}
\caption{Impact of varying scaling factors $\alpha$ and $\beta$ on performance and profiling time of sigmoid approximation on \textbf{CV16} and \textbf{Xsum}. 
The values are computed on a random sample of 10\% of each dataset.}
\label{tab:scale_ablation_full}
\end{table}
\subsection{Relation to other optimization methods}
Our method is orthogonal to other optimizations of speculative decoding. 
Whenever speculative sampling is used, our kernel can serve as a drop-in replacement for the standard implementation. 
For example, our proposed method can be integrated with the recently proposed self-speculative decoding approach \cite{zhang2023draft}. 
Instead of using a separate draft model, self-speculative decoding samples draft tokens by skipping some layers of the target model. 
Afterwards, it follows the same draft verification and resampling procedure as the original speculative decoding, which can be further accelerated with our optimization method. 

Our method is also orthogonal to other approaches for accelerating decoding. 
For instance, FlashAttention \cite{dao2022flashattention,dao2024flashattention2}, which focuses on optimizing the attention computation, can be easily combined with our method to further improve efficiency. 
\subsection{Overhead caused by resampling}
To assess the overhead caused by resampling, we followed \citet{chen2023sampling} and computed average acceptance rates of draft tokens for various models on 10\% of Xsum. 
\autoref{tab:accept_rates} includes the acceptance rates using a varying number of draft tokens ($\gamma \in \lbrace 3,5,10,15 \rbrace$), as well as the average execution time per decoding step of the speculative sampling algorithm. 
Since our exact optimization method aims to generate the same tokens as the baseline implementation, we expect the acceptance rates to be the same. 
\autoref{tab:accept_rates} shows that this is indeed the case for all choices of $\gamma$ and model combinations. 
\autoref{tab:accept_rates} also shows that the acceptance rates with the sigmoid optimization method are often higher than the acceptance rates of the baseline and the exact method. 
However, these higher acceptance rates, do not have a significant effect on the average execution time. 
In particular, the acceptance rates of the three methods (sigmoid, exact, and baseline) are similar for the Qwen model combination ($\sim$48\% with $\gamma = 10$), but the sigmoid approximation achieves better execution times than the exact optimization and the baseline. 
The average execution times provided \autoref{tab:accept_rates} are consistent with the ones presented in \autoref{fig:mean_sampling_time-text}. 
\begin{table*}[h!]
\centering
\resizebox{1.0\textwidth}{!}{
\begin{tabular}{c|c|c|cccc|cccc}
\toprule
\multirow{2}{*}{\textbf{Target Model} }  & \multirow{2}{*}{\textbf{Draft Model} }  & \multirow{2}{*}{\textbf{Method} }  & \multicolumn{4}{c|}{\textbf{Acceptance Rate}} & \multicolumn{4}{c}{\textbf{Average Execution Time}} \\
&  &  & $\gamma = 3$ & $\gamma = 5$ & $\gamma = 10$ & $\gamma = 15$ & $\gamma = 3$ & $\gamma = 5$ & $\gamma = 10$ & $\gamma = 15$ \\
\midrule
\multirow{3}{*}{Gemma 7B} & \multirow{3}{*}{Gemma 2B} & Sigmoid & 56.2\% & 57.0\% & 57.1\% & 56.1\% & 2.95 ms & 3.10 ms & 3.09 ms & 3.12 ms \\
 & & Exact   & 48.5\% & 48.4\% & 48.6\% & 46.4\% & 4.35 ms & 4.36 ms & 4.40 ms & 4.40 ms \\
&  & Baseline & 48.5\% & 48.4\% & 48.6\% & 46.4\% & 4.86 ms & 4.88 ms & 4.90 ms & 4.96 ms \\
\hline
\multirow{3}{*}{Qwen 7B} & \multirow{3}{*}{Qwen 0.5B} & Sigmoid & 46.2\% & 47.8\% & 48.1\% & 47.8\% & 2.92 ms & 2.89 ms & 2.84 ms & 2.83 ms \\
 & & Exact   & 46.8\% & 45.7\% & 48.4\% & 48.0\% & 3.91 ms & 3.97 ms & 3.88 ms & 3.95 ms \\
 & & Baseline & 46.8\% & 45.7\% & 48.4\% & 48.0\% & 4.45 ms & 4.49 ms & 4.41 ms & 4.48 ms \\
\hline
\multirow{3}{*}{Llama2 7B} & \multirow{3}{*}{\makecell{Sheared \\ Llama 1.3B}} & Sigmoid & 58.4\% & 59.3\% & 58.1\% & 59.4\% & 3.13 ms & 3.26 ms & 3.20 ms & 3.52 ms \\
& & Exact   & 51.5\% & 53.7\% & 56.2\% & 54.4\% & 3.41 ms & 3.42 ms & 3.49 ms & 3.78 ms \\
& & Baseline & 51.5\% & 53.7\% & 56.2\% & 54.4\% & 3.96 ms & 3.95 ms & 3.96 ms & 4.31 ms \\
\bottomrule
\end{tabular}
}
\vspace{-0.1in}
\caption{Comparison of acceptance rates by optimization type for different models on a random sample of
10\% of the Xsum dataset with varying $\gamma$.}
\label{tab:accept_rates}
\end{table*}

%% file: main.bbl
\begin{thebibliography}{76}
\providecommand{\natexlab}[1]{#1}

\bibitem[{Achiam et~al.(2024)}]{openai2024gpt4}
Josh Achiam et~al. 2024.
\newblock \href {https://arxiv.org/abs/2303.08774} {{GPT}-4 technical report}.
\newblock \emph{Preprint}, arXiv:2303.08774.

\bibitem[{Anil et~al.(2023)Anil, Borgeaud, Wu, Alayrac, Yu, Soricut, Schalkwyk, Dai, Hauth et~al.}]{geminiteam2024gemini}
Rohan Anil, Sebastian Borgeaud, Yonghui Wu, Jean-Baptiste Alayrac, Jiahui Yu, Radu Soricut, Johan Schalkwyk, Andrew~M Dai, Anja Hauth, et~al. 2023.
\newblock Gemini: a family of highly capable multimodal models.
\newblock \emph{arXiv preprint arXiv:2312.11805}.

\bibitem[{Ardila et~al.(2020)Ardila, Branson, Davis, Kohler, Meyer, Henretty, Morais, Saunders, Tyers, and Weber}]{ardila2020cv}
Rosana Ardila, Megan Branson, Kelly Davis, Michael Kohler, Josh Meyer, Michael Henretty, Reuben Morais, Lindsay Saunders, Francis Tyers, and Gregor Weber. 2020.
\newblock {Common Voice}: A massively-multilingual speech corpus.
\newblock In \emph{LREC}, pages 4218--4222.

\bibitem[{Bai et~al.(2023{\natexlab{a}})Bai, Bai, Chu, Cui, Dang, Deng, Fan, Ge, Han, Huang et~al.}]{bai2023qwen}
Jinze Bai, Shuai Bai, Yunfei Chu, Zeyu Cui, Kai Dang, Xiaodong Deng, Yang Fan, Wenbin Ge, Yu~Han, Fei Huang, et~al. 2023{\natexlab{a}}.
\newblock Qwen technical report.
\newblock \emph{arXiv preprint arXiv:2309.16609}.

\bibitem[{Bai et~al.(2023{\natexlab{b}})Bai, Chen, Wang, Xiong, and Mei}]{bai2023transformers}
Yu~Bai, Fan Chen, Huan Wang, Caiming Xiong, and Song Mei. 2023{\natexlab{b}}.
\newblock Transformers as statisticians: Provable in-context learning with in-context algorithm selection.
\newblock \emph{Advances in Neural Information Processing Systems (Neurips)}.

\bibitem[{Bondarenko et~al.(2023)Bondarenko, Nagel, and Blankevoort}]{bondarenko2023quantizable}
Yelysei Bondarenko, Markus Nagel, and Tijmen Blankevoort. 2023.
\newblock Quantizable transformers: Removing outliers by helping attention heads do nothing.
\newblock \emph{Advances in Neural Information Processing Systems (NeurIPS)}.

\bibitem[{Bridle(1989)}]{bridle89softmax}
John Bridle. 1989.
\newblock Training stochastic model recognition algorithms as networks can lead to maximum mutual information estimation of parameters.
\newblock \emph{Advances in Neural Information Processing Systems (NeurIPS)}.

\bibitem[{Cai et~al.(2024)Cai, Li, Geng, Peng, Lee, Chen, and Dao}]{cai2024medusa}
Tianle Cai, Yuhong Li, Zhengyang Geng, Hongwu Peng, Jason~D Lee, Deming Chen, and Tri Dao. 2024.
\newblock Medusa: Simple {LLM} inference acceleration framework with multiple decoding heads.
\newblock \emph{arXiv preprint arXiv:2401.10774}.

\bibitem[{Chen et~al.(2023{\natexlab{a}})Chen, Borgeaud, Irving, Lespiau, Sifre, and Jumper}]{chen2023sampling}
Charlie Chen, Sebastian Borgeaud, Geoffrey Irving, Jean-Baptiste Lespiau, Laurent Sifre, and John Jumper. 2023{\natexlab{a}}.
\newblock Accelerating large language model decoding with speculative sampling.
\newblock \emph{arXiv preprint arXiv:2302.01318}.

\bibitem[{Chen et~al.(2023{\natexlab{b}})Chen, Gao, Waris, Liu, and Lombardi}]{ke2023approx}
Ke~Chen, Yue Gao, Haroon Waris, Weiqiang Liu, and Fabrizio Lombardi. 2023{\natexlab{b}}.
\newblock Approximate softmax functions for energy-efficient deep neural networks.
\newblock \emph{IEEE Transactions on Very Large Scale Integration (VLSI) Systems}.

\bibitem[{Cheng et~al.(2014)Cheng, Grossman, and McKercher}]{cheng2014professional}
J.~Cheng, M.~Grossman, and T.~McKercher. 2014.
\newblock \emph{Professional CUDA C Programming}.
\newblock Wiley.

\bibitem[{Chi et~al.(2023)Chi, Kim, Hickmann, Li, Chi, Atchariyachanvanit, Yu, Chi, Dai, Rammoorthy et~al.}]{chi2023distillery}
Ryan~A Chi, Jeremy Kim, Scott Hickmann, Siyan Li, Gordon Chi, Thanawan Atchariyachanvanit, Katherine Yu, Nathan~A Chi, Gary Dai, Shashank Rammoorthy, et~al. 2023.
\newblock Dialogue distillery: Crafting interpolable, interpretable, and introspectable dialogue from {LLMs}.
\newblock \emph{Alexa Prize SocialBot Grand Challenge}, 5.

\bibitem[{Dao(2024)}]{dao2024flashattention2}
Tri Dao. 2024.
\newblock {FlashAttention-2}: Faster attention with better parallelism and work partitioning.
\newblock \emph{International Conference on Learning Representations {(ICLR)}}.

\bibitem[{Dao et~al.(2022)Dao, Fu, Ermon, Rudra, and Re}]{dao2022flashattention}
Tri Dao, Daniel~Y Fu, Stefano Ermon, Atri Rudra, and Christopher Re. 2022.
\newblock Flashattention: Fast and memory-efficient exact attention with {IO}-awareness.
\newblock \emph{Neural Information Processing Systems ({NeurIPS})}.

\bibitem[{Dettmers et~al.(2022)Dettmers, Lewis, Belkada, and Zettlemoyer}]{dettmers2022gptint}
Tim Dettmers, Mike Lewis, Younes Belkada, and Luke Zettlemoyer. 2022.
\newblock {GPT}3.int8(): 8-bit matrix multiplication for transformers at scale.
\newblock In \emph{Advances in Neural Information Processing Systems (NeurIPS)}.

\bibitem[{Fernandez et~al.(2023)Fernandez, Kahn, Na, Bisk, and Strubell}]{fernandez2023frameworktax}
Jared Fernandez, Jacob Kahn, Clara Na, Yonatan Bisk, and Emma Strubell. 2023.
\newblock \href {https://doi.org/10.18653/v1/2023.emnlp-main.98} {The framework tax: Disparities between inference efficiency in {NLP} research and deployment}.
\newblock In \emph{Proceedings of the 2023 Conference on Empirical Methods in Natural Language Processing}, pages 1588--1600, Singapore. Association for Computational Linguistics.

\bibitem[{Fischer et~al.(2024)Fischer, Gemmell, Tecklenburg, Mackie, Rossetto, and Dalton}]{fischer2024grillbot}
Sophie Fischer, Carlos Gemmell, Niklas Tecklenburg, Iain Mackie, Federico Rossetto, and Jeffrey Dalton. 2024.
\newblock {GRILLBot} in practice: Lessons and tradeoffs deploying large language models for adaptable conversational task assistants.
\newblock \emph{arXiv preprint arXiv:2402.07647}.

\bibitem[{Gandhi et~al.(2023)Gandhi, von Platen, and Rush}]{gandhi2023distilwhisper}
Sanchit Gandhi, Patrick von Platen, and Alexander~M Rush. 2023.
\newblock Distil-whisper: Robust knowledge distillation via large-scale pseudo labelling.
\newblock \emph{arXiv preprint arXiv:2311.00430}.

\bibitem[{Geng et~al.(2018)Geng, Lin, Zhao, Kong, Aly, and Chandrasekhar}]{Geng2018HardwareAwareSA}
Xue Geng, Jie Lin, Bin Zhao, Anmin Kong, Mohamed M.~Sabry Aly, and Vijay~Ramaseshan Chandrasekhar. 2018.
\newblock Hardware-aware softmax approximation for deep neural networks.
\newblock \emph{Asian Conference on Computer Vision}.

\bibitem[{Grave et~al.(2017)Grave, Joulin, Ciss\'{e}, Facebook AI~Research, and J\'{e}gou}]{grave2017adaptivesoftmax}
\'{E}douard Grave, Armand Joulin, Moustapha Ciss\'{e}, David~Grangier Facebook AI~Research, and Herv\'{e} J\'{e}gou. 2017.
\newblock Efficient softmax approximation for {GPUs}.
\newblock \emph{International Conference on Machine Learning (ICML)}.

\bibitem[{Gromov et~al.(2024)Gromov, Tirumala, Shapourian, Glorioso, and Roberts}]{gromov2024unreasonable}
Andrey Gromov, Kushal Tirumala, Hassan Shapourian, Paolo Glorioso, and Daniel~A Roberts. 2024.
\newblock The unreasonable ineffectiveness of the deeper layers.
\newblock \emph{arXiv preprint arXiv:2403.17887}.

\bibitem[{Harris(2007)}]{harris2007parallel_reduce}
Mark Harris. 2007.
\newblock \href {https://developer.download.nvidia.com/assets/cuda/files/reduction.pdf} {Optimizing parallel reduction in cuda}.
\newblock Technical report, NVIDIA.
\newblock Accessed: 2024-06-12.

\bibitem[{Hinton et~al.(2015)Hinton, Vinyals, and Dean}]{hinton15distill}
Geoffrey Hinton, Oriol Vinyals, and Jeffrey Dean. 2015.
\newblock Distilling the knowledge in a neural network.
\newblock In \emph{NIPS Deep Learning and Representation Learning Workshop}.

\bibitem[{Hron et~al.(2020)Hron, Bahri, Sohl-Dickstein, and Novak}]{Hron2020InfiniteAN}
Jiri Hron, Yasaman Bahri, Jascha~Narain Sohl-Dickstein, and Roman Novak. 2020.
\newblock Infinite attention: Nngp and ntk for deep attention networks.
\newblock \emph{International Conference on Machine Learning (ICML)}.

\bibitem[{Hsieh et~al.(2023)Hsieh, Li, Yeh, Nakhost, Fujii, Ratner, Krishna, Lee, and Pfister}]{hsieh-etal-2023-distilling}
Cheng-Yu Hsieh, Chun-Liang Li, Chih-kuan Yeh, Hootan Nakhost, Yasuhisa Fujii, Alex Ratner, Ranjay Krishna, Chen-Yu Lee, and Tomas Pfister. 2023.
\newblock \href {https://doi.org/10.18653/v1/2023.findings-acl.507} {Distilling step-by-step! outperforming larger language models with less training data and smaller model sizes}.
\newblock In \emph{Findings of the Association for Computational Linguistics: ACL 2023}, pages 8003--8017, Toronto, Canada. Association for Computational Linguistics.

\bibitem[{Hsu et~al.(2021)Hsu, Bolte, Tsai, Lakhotia, Salakhutdinov, and Mohamed}]{weining2021hubert}
Wei-Ning Hsu, Benjamin Bolte, Yao-Hung~Hubert Tsai, Kushal Lakhotia, Ruslan Salakhutdinov, and Abdelrahman Mohamed. 2021.
\newblock Hubert: Self-supervised speech representation learning by masked prediction of hidden units.
\newblock \emph{IEEE/ACM Trans. Audio, Speech and Lang. Proc.}, page 3451–3460.

\bibitem[{Hua et~al.(2022)Hua, Dai, Liu, and Le}]{hua2022quality}
Weizhe Hua, Zihang Dai, Hanxiao Liu, and Quoc Le. 2022.
\newblock Transformer quality in linear time.
\newblock \emph{International Conference on Machine Learning (ICML)}.

\bibitem[{Jia et~al.(2018)Jia, Maggioni, Staiger, and Scarpazza}]{jia2018dissecting}
Zhe Jia, Marco Maggioni, Benjamin Staiger, and Daniele~P. Scarpazza. 2018.
\newblock Dissecting the {NVIDIA} volta {GPU} architecture via microbenchmarking.
\newblock \emph{arxiv preprint arXiv:1804.06826}.

\bibitem[{Jiao et~al.(2020)Jiao, Yin, Shang, Jiang, Chen, Li, Wang, and Liu}]{jiao2020tinybert}
Xiaoqi Jiao, Yichun Yin, Lifeng Shang, Xin Jiang, Xiao Chen, Linlin Li, Fang Wang, and Qun Liu. 2020.
\newblock \href {https://doi.org/10.18653/v1/2020.findings-emnlp.372} {{T}iny{BERT}: Distilling {BERT} for natural language understanding}.
\newblock In \emph{Findings of the Association for Computational Linguistics: EMNLP 2020}, pages 4163--4174, Online. Association for Computational Linguistics.

\bibitem[{Lagunas et~al.(2021)Lagunas, Charlaix, Sanh, and Rush}]{lagunas2021blockpruning}
Fran{\c{c}}ois Lagunas, Ella Charlaix, Victor Sanh, and Alexander Rush. 2021.
\newblock \href {https://doi.org/10.18653/v1/2021.emnlp-main.829} {Block pruning for faster transformers}.
\newblock In \emph{Proceedings of the 2021 Conference on Empirical Methods in Natural Language Processing}, pages 10619--10629, Online and Punta Cana, Dominican Republic. Association for Computational Linguistics.

\bibitem[{Lam et~al.(1991)Lam, Rothberg, and Wolf}]{lam1991tiling}
Monica~D. Lam, Edward~E. Rothberg, and Michael~E. Wolf. 1991.
\newblock The cache performance and optimizations of blocked algorithms.
\newblock \emph{Architectural Support for Programming Languages and Operating Systems (ASPLOS)}.

\bibitem[{Lee and Lee(2023)}]{lee2023softmax}
Changhyeon Lee and Seulki Lee. 2023.
\newblock Softmax output approximation for activation memory-efficient training of attention-based networks.
\newblock \emph{Advances in Neural Information Processing Systems (NeurIPS)}, 36.

\bibitem[{Leviathan et~al.(2023)Leviathan, Kalman, and Matias}]{leviathan2023decoding}
Yaniv Leviathan, Matan Kalman, and Yossi Matias. 2023.
\newblock Fast inference from transformers via speculative decoding.
\newblock \emph{International Conference on Machine Learning (ICML)}.

\bibitem[{Li et~al.(2022)Li, Bhojanapalli, Zaheer, Reddi, and Kumar}]{li2022robust}
Zhiyuan Li, Srinadh Bhojanapalli, Manzil Zaheer, Sashank Reddi, and Sanjiv Kumar. 2022.
\newblock Robust training of neural networks using scale invariant architectures.
\newblock \emph{International Conference on Machine Learning (ICML)}.

\bibitem[{Lin(2004)}]{lin-2004-rouge}
Chin-Yew Lin. 2004.
\newblock \href {https://aclanthology.org/W04-1013} {{ROUGE}: A package for automatic evaluation of summaries}.
\newblock In \emph{Text Summarization Branches Out}, pages 74--81, Barcelona, Spain. Association for Computational Linguistics.

\bibitem[{Mesnard et~al.(2024)Mesnard, Hardin, Dadashi, Bhupatiraju, Pathak, Sifre, Rivi{\`e}re, Kale, Love et~al.}]{gemmateam2024gemma}
Thomas Mesnard, Cassidy Hardin, Robert Dadashi, Surya Bhupatiraju, Shreya Pathak, Laurent Sifre, Morgane Rivi{\`e}re, Mihir~Sanjay Kale, Juliette Love, et~al. 2024.
\newblock Gemma: Open models based on gemini research and technology.
\newblock \emph{arXiv preprint arXiv:2403.08295}.

\bibitem[{Miao et~al.(2024)Miao, Oliaro, Zhang, Cheng, Wang, Zhang, Wong, Zhu, Yang, Shi, Shi, Chen, Arfeen, Abhyankar, and Jia}]{miao2024tokentree}
Xupeng Miao, Gabriele Oliaro, Zhihao Zhang, Xinhao Cheng, Zeyu Wang, Zhengxin Zhang, Rae Ying~Yee Wong, Alan Zhu, Lijie Yang, Xiaoxiang Shi, Chunan Shi, Zhuoming Chen, Daiyaan Arfeen, Reyna Abhyankar, and Zhihao Jia. 2024.
\newblock Specinfer: Accelerating large language model serving with tree-based speculative inference and verification.
\newblock In \emph{Proceedings of the 29th ACM International Conference on Architectural Support for Programming Languages and Operating Systems, Volume 3}.

\bibitem[{Milakov and Gimelshein(2018)}]{milakov2018online}
Maxim Milakov and Natalia Gimelshein. 2018.
\newblock Online normalizer calculation for softmax.
\newblock \emph{arXiv preprint arXiv:1805.02867}.

\bibitem[{Monea et~al.(2023)Monea, Joulin, and Grave}]{monea2023pass}
Giovanni Monea, Armand Joulin, and Edouard Grave. 2023.
\newblock {PaSS}: Parallel speculative sampling.
\newblock \emph{arXiv preprint arXiv:2311.13581}.

\bibitem[{Nagel et~al.(2021)Nagel, Fournarakis, Amjad, Bondarenko, Van~Baalen, and Blankevoort}]{nagel2021white}
Markus Nagel, Marios Fournarakis, Rana~Ali Amjad, Yelysei Bondarenko, Mart Van~Baalen, and Tijmen Blankevoort. 2021.
\newblock A white paper on neural network quantization.
\newblock \emph{arXiv preprint arXiv:2106.08295}.

\bibitem[{Nallapati et~al.(2016)Nallapati, Zhou, dos Santos, Gul{\c{c}}ehre, and Xiang}]{nallapati2016cnn}
Ramesh Nallapati, Bowen Zhou, Cicero dos Santos, {\c{C}}a{\u{g}}lar Gul{\c{c}}ehre, and Bing Xiang. 2016.
\newblock \href {https://doi.org/10.18653/v1/K16-1028} {Abstractive text summarization using sequence-to-sequence {RNN}s and beyond}.
\newblock In \emph{Proceedings of the 20th {SIGNLL} Conference on Computational Natural Language Learning}, pages 280--290, Berlin, Germany. Association for Computational Linguistics.

\bibitem[{Narayan et~al.(2018)Narayan, Cohen, and Lapata}]{narayan2018xsum}
Shashi Narayan, Shay~B. Cohen, and Mirella Lapata. 2018.
\newblock \href {https://doi.org/10.18653/v1/D18-1206} {Don{'}t give me the details, just the summary! topic-aware convolutional neural networks for extreme summarization}.
\newblock In \emph{Proceedings of the 2018 Conference on Empirical Methods in Natural Language Processing}, pages 1797--1807, Brussels, Belgium. Association for Computational Linguistics.

\bibitem[{Neal(2003)}]{neal2003rejectionsampling}
Radford~M. Neal. 2003.
\newblock \href {https://doi.org/10.1214/aos/1056562461} {Slice sampling}.
\newblock \emph{The Annals of Statistics}, 31(3).

\bibitem[{{NVIDIA Corporation}(2020)}]{nvidia_a100_2020}
{NVIDIA Corporation}. 2020.
\newblock \href {https://resources.nvidia.com/en-us-genomics-ep/ampere-architecture-white-paper} {{NVIDIA A100} tensor core {GPU} architecture}.
\newblock Technical report, NVIDIA Corporation.

\bibitem[{Panayotov et~al.(2015)Panayotov, Chen, Povey, and Khudanpur}]{panayotov2015librispeech}
Vassil Panayotov, Guoguo Chen, Daniel Povey, and Sanjeev Khudanpur. 2015.
\newblock Librispeech: An asr corpus based on public domain audio books.
\newblock In \emph{ICASSP}.

\bibitem[{Paszke et~al.(2019)Paszke, Gross, Massa, Lerer, Bradbury, Chanan, Killeen, Lin, Gimelshein, Antiga, Desmaison, Kopf, Yang, DeVito, Raison, Tejani, Chilamkurthy, Steiner, Fang, Bai, and Chintala}]{paszke2019pytorch}
Adam Paszke, Sam Gross, Francisco Massa, Adam Lerer, James Bradbury, Gregory Chanan, Trevor Killeen, Zeming Lin, Natalia Gimelshein, Luca Antiga, Alban Desmaison, Andreas Kopf, Edward Yang, Zachary DeVito, Martin Raison, Alykhan Tejani, Sasank Chilamkurthy, Benoit Steiner, Lu~Fang, Junjie Bai, and Soumith Chintala. 2019.
\newblock Pytorch: An imperative style, high-performance deep learning library.
\newblock \emph{Neural Information Processing Systems {(NeurIPS)}}.

\bibitem[{Pope et~al.(2023)Pope, Douglas, Chowdhery, Devlin, Bradbury, Heek, Xiao, Agrawal, and Dean}]{pope2022efficiently}
Reiner Pope, Sholto Douglas, Aakanksha Chowdhery, Jacob Devlin, James Bradbury, Jonathan Heek, Kefan Xiao, Shivani Agrawal, and Jeff Dean. 2023.
\newblock Efficiently scaling transformer inference.
\newblock \emph{Machine Learning and Systems}.

\bibitem[{Rabe and Staats(2021)}]{rabe2022selfattention}
Markus~N Rabe and Charles Staats. 2021.
\newblock Self-attention does not need $o(n^2)$ memory.
\newblock \emph{arXiv preprint arXiv:2112.05682}.

\bibitem[{Radford et~al.(2022)Radford, Kim, Xu, Brockman, McLeavey, and Sutskever}]{radford2022whisper}
Alec Radford, Jong~Wook Kim, Tao Xu, Greg Brockman, Christine McLeavey, and Ilya Sutskever. 2022.
\newblock \href {https://arxiv.org/abs/2212.04356} {Robust speech recognition via large-scale weak supervision}.
\newblock \emph{Preprint}, arXiv:2212.04356.
\newblock ArXiv:2212.04356.

\bibitem[{Ramapuram et~al.(2024)Ramapuram, Danieli, Dhekane, Weers, Busbridge, Ablin, Likhomanenko, Digani, Gu, Shidani, and Webb}]{ramapuram2024sigmoidatt}
Jason Ramapuram, Federico Danieli, Eeshan Dhekane, Floris Weers, Dan Busbridge, Pierre Ablin, Tatiana Likhomanenko, Jagrit Digani, Zijin Gu, Amitis Shidani, and Russ Webb. 2024.
\newblock \href {https://arxiv.org/abs/2409.04431} {Theory, analysis, and best practices for sigmoid self-attention}.
\newblock \emph{Preprint}, arXiv:2409.04431.

\bibitem[{Rousseau et~al.(2012)Rousseau, Del{\'e}glise, and Est{\`e}ve}]{rousseau2012ted}
Anthony Rousseau, Paul Del{\'e}glise, and Yannick Est{\`e}ve. 2012.
\newblock {TED}-{LIUM}: an automatic speech recognition dedicated corpus.
\newblock In \emph{LREC}, pages 125--129.

\bibitem[{Ryoo et~al.(2008)Ryoo, Rodrigues, Stone, Baghsorkhi, Ueng, Stratton, and Hwu}]{ryoo2008optimization}
Shane Ryoo, Christopher~I Rodrigues, Sam~S Stone, Sara~S Baghsorkhi, Sain-Zee Ueng, John~A Stratton, and Wen-mei~W Hwu. 2008.
\newblock Program optimization space pruning for a multithreaded gpu.
\newblock In \emph{IEEE/ACM international symposium on Code generation and optimization}.

\bibitem[{Sanh et~al.(2019)Sanh, Debut, Chaumond, and Wolf}]{sanh2020distilbert}
Victor Sanh, Lysandre Debut, Julien Chaumond, and Thomas Wolf. 2019.
\newblock Distilbert, a distilled version of bert: smaller, faster, cheaper and lighter.
\newblock \emph{arXiv preprint arXiv:1910.01108}.

\bibitem[{Shazeer(2019)}]{shazeer2019fast}
Noam Shazeer. 2019.
\newblock Fast transformer decoding: One write-head is all you need.
\newblock \emph{arXiv preprint arXiv:1911.02150}.

\bibitem[{Shazeer(2020)}]{shazeer2020glu}
Noam Shazeer. 2020.
\newblock {GLU} variants improve transformer.
\newblock \emph{arXiv preprint arXiv:2002.05202}.

\bibitem[{Shen et~al.(2023)Shen, Guo, Tan, Tang, Wang, and Bian}]{shen2023study}
Kai Shen, Junliang Guo, Xu~Tan, Siliang Tang, Rui Wang, and Jiang Bian. 2023.
\newblock A study on relu and softmax in transformer.
\newblock \emph{arXiv preprint arXiv:2302.06461}.

\bibitem[{Shim et~al.(2017)Shim, Lee, Choi, Boo, and Sung}]{shim2017svdsoftmax}
Kyuhong Shim, Minjae Lee, Iksoo Choi, Yoonho Boo, and Wonyong Sung. 2017.
\newblock Svd-softmax: Fast softmax approximation on large vocabulary neural networks.
\newblock \emph{Advances in Neural Information Processing Systems (NeurIPS)}.

\bibitem[{Spector and Re(2023)}]{spector2023accelerating}
Benjamin Spector and Chris Re. 2023.
\newblock \href {https://arxiv.org/abs/2308.04623} {Accelerating {LLM} inference with staged speculative decoding}.
\newblock \emph{arXiv preprint arXiv:2308.04623}.

\bibitem[{Stern et~al.(2018)Stern, Shazeer, and Uszkoreit}]{stern2018specdec}
Mitchell Stern, Noam Shazeer, and Jakob Uszkoreit. 2018.
\newblock Blockwise parallel decoding for deep autoregressive models.
\newblock \emph{Advances in Neural Information Processing Systems (NeurIPS)}.

\bibitem[{Stock et~al.(2021)Stock, Fan, Graham, Grave, Gribonval, Jegou, and Joulin}]{stock2021training}
Pierre Stock, Angela Fan, Benjamin Graham, Edouard Grave, R{\'e}mi Gribonval, Herve Jegou, and Armand Joulin. 2021.
\newblock Training with quantization noise for extreme model compression.
\newblock In \emph{International Conference on Learning Representations (ICLR)}.

\bibitem[{Su et~al.(2024)Su, Ahmed, Lu, Pan, Bo, and Liu}]{su2023roformer}
Jianlin Su, Murtadha Ahmed, Yu~Lu, Shengfeng Pan, Wen Bo, and Yunfeng Liu. 2024.
\newblock Roformer: Enhanced transformer with rotary position embedding.
\newblock \emph{Neurocomputing}, 568:127063.

\bibitem[{Sun et~al.(2019)Sun, Cheng, Gan, and Liu}]{sun2019patient}
Siqi Sun, Yu~Cheng, Zhe Gan, and Jingjing Liu. 2019.
\newblock \href {https://doi.org/10.18653/v1/D19-1441} {Patient knowledge distillation for {BERT} model compression}.
\newblock In \emph{Proceedings of the 2019 Conference on Empirical Methods in Natural Language Processing and the 9th International Joint Conference on Natural Language Processing (EMNLP-IJCNLP)}, pages 4323--4332, Hong Kong, China. Association for Computational Linguistics.

\bibitem[{Touvron et~al.(2023{\natexlab{a}})Touvron, Lavril, Izacard, Martinet, Lachaux, Lacroix, Rozi{\`e}re, Goyal, Hambro, Azhar et~al.}]{touvron2023llama}
Hugo Touvron, Thibaut Lavril, Gautier Izacard, Xavier Martinet, Marie-Anne Lachaux, Timoth{\'e}e Lacroix, Baptiste Rozi{\`e}re, Naman Goyal, Eric Hambro, Faisal Azhar, et~al. 2023{\natexlab{a}}.
\newblock Llama: Open and efficient foundation language models.
\newblock \emph{arXiv preprint arXiv:2302.13971}.

\bibitem[{Touvron et~al.(2023{\natexlab{b}})}]{touvron2023llama2}
Hugo Touvron et~al. 2023{\natexlab{b}}.
\newblock Llama 2: Open foundation and fine-tuned chat models.
\newblock \emph{arXiv preprint arXiv:2307.09288}.

\bibitem[{Vaswani et~al.(2017)Vaswani, Shazeer, Parmar, Uszkoreit, Jones, Gomez, Kaiser, and Polosukhin}]{vaswani17attention}
Ashish Vaswani, Noam Shazeer, Niki Parmar, Jakob Uszkoreit, Llion Jones, Aidan~N Gomez, {\L}ukasz Kaiser, and Illia Polosukhin. 2017.
\newblock Attention is all you need.
\newblock \emph{Advances in neural information processing systems (NeurIPS)}.

\bibitem[{Voita et~al.(2019)Voita, Talbot, Moiseev, Sennrich, and Titov}]{voita2019analyzing}
Elena Voita, David Talbot, Fedor Moiseev, Rico Sennrich, and Ivan Titov. 2019.
\newblock \href {https://doi.org/10.18653/v1/P19-1580} {Analyzing multi-head self-attention: Specialized heads do the heavy lifting, the rest can be pruned}.
\newblock In \emph{Proceedings of the 57th Annual Meeting of the Association for Computational Linguistics}, pages 5797--5808, Florence, Italy. Association for Computational Linguistics.

\bibitem[{Wolf et~al.(2020)Wolf, Debut, Sanh, Chaumond, Delangue, Moi, Cistac, Rault, Louf, Funtowicz, Davison, Shleifer, von Platen, Ma, Jernite, Plu, Xu, Le~Scao, Gugger, Drame, Lhoest, and Rush}]{wolf2020transformers}
Thomas Wolf, Lysandre Debut, Victor Sanh, Julien Chaumond, Clement Delangue, Anthony Moi, Pierric Cistac, Tim Rault, Remi Louf, Morgan Funtowicz, Joe Davison, Sam Shleifer, Patrick von Platen, Clara Ma, Yacine Jernite, Julien Plu, Canwen Xu, Teven Le~Scao, Sylvain Gugger, Mariama Drame, Quentin Lhoest, and Alexander Rush. 2020.
\newblock \href {https://doi.org/10.18653/v1/2020.emnlp-demos.6} {Transformers: State-of-the-art natural language processing}.
\newblock In \emph{Proceedings of the 2020 Conference on Empirical Methods in Natural Language Processing: System Demonstrations}, pages 38--45, Online. Association for Computational Linguistics.

\bibitem[{Wortsman et~al.(2023)Wortsman, Lee, Gilmer, and Kornblith}]{wortsman2023replacing}
Mitchell Wortsman, Jaehoon Lee, Justin Gilmer, and Simon Kornblith. 2023.
\newblock Replacing softmax with relu in vision transformers.
\newblock \emph{arXiv preprint arXiv:2309.08586}.

\bibitem[{Xia et~al.(2023)Xia, Ge, Wang, Chen, Wei, and Sui}]{xia-etal-2023-speculative}
Heming Xia, Tao Ge, Peiyi Wang, Si-Qing Chen, Furu Wei, and Zhifang Sui. 2023.
\newblock \href {https://doi.org/10.18653/v1/2023.findings-emnlp.257} {Speculative decoding: Exploiting speculative execution for accelerating seq2seq generation}.
\newblock In \emph{Findings of the Association for Computational Linguistics: EMNLP 2023}, pages 3909--3925, Singapore. Association for Computational Linguistics.

\bibitem[{Xia et~al.(2024{\natexlab{a}})Xia, Yang, Dong, Wang, Li, Ge, Liu, Li, and Sui}]{xia2024unlocking}
Heming Xia, Zhe Yang, Qingxiu Dong, Peiyi Wang, Yongqi Li, Tao Ge, Tianyu Liu, Wenjie Li, and Zhifang Sui. 2024{\natexlab{a}}.
\newblock Unlocking efficiency in large language model inference: A comprehensive survey of speculative decoding.
\newblock \emph{arXiv preprint arXiv:2401.07851}.

\bibitem[{Xia et~al.(2024{\natexlab{b}})Xia, Gao, Zeng, and Chen}]{xia2024sheared}
Mengzhou Xia, Tianyu Gao, Zhiyuan Zeng, and Danqi Chen. 2024{\natexlab{b}}.
\newblock Sheared {LL}a{MA}: Accelerating language model pre-training via structured pruning.
\newblock \emph{International Conference on Learning Representations (ICLR)}.

\bibitem[{Yang et~al.(2024)Yang, Huang, Dai, and Chen}]{yang2024multicandidate}
Sen Yang, Shujian Huang, Xinyu Dai, and Jiajun Chen. 2024.
\newblock Multi-candidate speculative decoding.
\newblock \emph{arXiv preprint arXiv:2401.06706}.

\bibitem[{Zhang et~al.(2024{\natexlab{a}})Zhang, Wang, Wang, Zhang, and Cheng}]{zhang2024recurrent}
Aonan Zhang, Chong Wang, Yi~Wang, Xuanyu Zhang, and Yunfei Cheng. 2024{\natexlab{a}}.
\newblock \href {https://arxiv.org/abs/2403.09919} {Recurrent drafter for fast speculative decoding in large language models}.
\newblock \emph{Preprint}, arXiv:2403.09919.

\bibitem[{Zhang and Sennrich(2019)}]{zhang2019rmsnorm}
Biao Zhang and Rico Sennrich. 2019.
\newblock Root mean square layer normalization.
\newblock \emph{Neural Information Processing Systems {(NeurIPS)}}.

\bibitem[{Zhang et~al.(2024{\natexlab{b}})Zhang, Wang, Li, Shou, Chen, Chen, and Mehrotra}]{zhang2023draft}
Jun Zhang, Jue Wang, Huan Li, Lidan Shou, Ke~Chen, Gang Chen, and Sharad Mehrotra. 2024{\natexlab{b}}.
\newblock \href {https://aclanthology.org/2024.acl-long.607} {Draft{\&} verify: Lossless large language model acceleration via self-speculative decoding}.
\newblock In \emph{Proceedings of the 62nd Annual Meeting of the Association for Computational Linguistics (Volume 1: Long Papers)}, pages 11263--11282, Bangkok, Thailand. Association for Computational Linguistics.

\bibitem[{Zhou et~al.(2024)Zhou, Lyu, Rawat, Menon, Rostamizadeh, Kumar, Kagy, and Agarwal}]{zhou2024distillspec}
Yongchao Zhou, Kaifeng Lyu, Ankit~Singh Rawat, Aditya~Krishna Menon, Afshin Rostamizadeh, Sanjiv Kumar, Jean-Fran{\c{c}}ois Kagy, and Rishabh Agarwal. 2024.
\newblock Distillspec: Improving speculative decoding via knowledge distillation.
\newblock In \emph{International Conference on Learning Representations (ICLR)}.

\end{thebibliography}
